\documentclass{article}

\usepackage{CJKutf8}
\usepackage{longcat_style}
\usepackage{adjustbox}
\usepackage[utf8]{inputenc} %
\usepackage[T1]{fontenc}    %
\usepackage{newunicodechar}
\usepackage{hyperref}       %
\usepackage{xcolor}
\usepackage[normalem]{ulem} %
\hypersetup{
    colorlinks=true,      %
    linkcolor=blue,      %
    urlcolor=blue,       %
    citecolor=blue,      %
    linkbordercolor=blue, %
    urlbordercolor=blue,
    citebordercolor=blue,
    pdfborderstyle={/S/U/W 1}, %
}
\usepackage{float}
\usepackage{url}            %
\usepackage{booktabs}       %
\usepackage{amsfonts}       %
\usepackage{nicefrac}       %
\usepackage{microtype}      %
\usepackage{lipsum}		%
\usepackage{graphicx}
\usepackage{natbib}
\usepackage{doi}
\usepackage{amsmath}
\usepackage{amssymb} %
\usepackage{xspace}
\usepackage{enumitem}
\usepackage{multirow}
\usepackage{subcaption} 
\usepackage{makecell}
\usepackage{hyperref, cleveref}
\usepackage{pifont}
\usepackage[inkscapelatex=false]{svg}
\usepackage{caption}
\DeclareCaptionLabelSeparator{pipe}{ | }
\usepackage{tcolorbox}
\usepackage{xcolor}
\usepackage[ruled,vlined]{algorithm2e}
\usepackage{amsmath, amssymb}
\newtcolorbox{coloredquote}[1][]{
    colback=green!5!white,  
    colframe=green!70!black, 
    boxrule=2pt,
    arc=7pt,
    left=6pt,
    right=6pt,
    top=4pt,
    bottom=4pt,
    title=#1
}

\setlist[itemize]{leftmargin=*}
\setlist[enumerate]{leftmargin=*}
\setlist[description]{leftmargin=*}

\title{What Transfers from Text to Vision? Capability Scaling Laws and Transfer Dynamics for VLMs}

\author{
Ziran Li\textsuperscript{1}\footnotemark[1], Qiang Wang\textsuperscript{2}\thanks{Equal Contribution.}, Zhengyu Chen\textsuperscript{1}, Shanglin Lei\textsuperscript{1} \\
\textbf{Borun Chen\textsuperscript{1}, Jingang Wang\textsuperscript{1}, Xunliang Cai\textsuperscript{1}}\\
\textsuperscript{1}Meituan \ \ \ \textsuperscript{2}Tsinghua University \\
	\texttt{\{liziran02,chenzhengyu04\}@meituan.com} \\
    \texttt{qiang-wa24@mails.tsinghua.edu.cn}\\
}

\renewcommand{\headeright}{\raisebox{-0.2\height}{\includegraphics[height=1.8em]{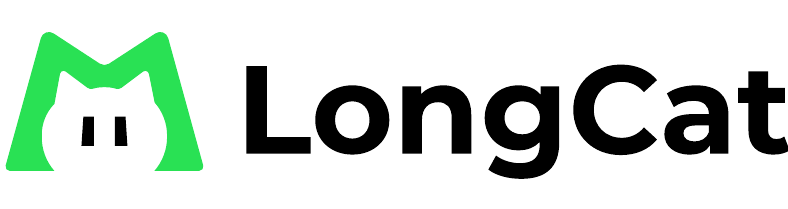}}}
\renewcommand{\shorttitle}{Capability-Driven Multimodal Scaling Law}

\begin{document}
\maketitle
\setcounter{footnote}{0}

\begin{abstract}
Choosing the right large language model (LLM) backbone is the most consequential decision when building a vision-language model (VLM), yet it remains fundamentally unprincipled: compute-based scaling laws fail to generalize across model families, and no framework exists for directly predicting VLM performance before training begins. We propose the \textbf{Capability-Driven Multimodal Scaling Law}, the first cross-family framework that predicts VLM benchmark accuracy from directly observable textual capability. Given a low-dimensional capability score $S$ extracted from LLM textual benchmarks via PCA, we model VLM performance as a function of $S$, with a per-backbone \emph{transfer rate} and an \emph{absorption rate} that quantifies data-scaling efficiency. To fit and validate the framework, we train over 150 VLMs on 34 LLMs spanning 7 model families under a strictly controlled recipe. Evaluations on more than 200 textual and 50 multimodal benchmarks show that the law accurately extrapolates transfer rate from models up to 8B parameters to 72B-scale backbones, predicts full VLM training trajectories with high fidelity, and generalizes to entirely held-out model families. Beyond the scaling law, our analysis surfaces actionable insights: certain textual benchmarks negatively correlate with multimodal performance, exposing latent benchmark-gaming behavior; base LLMs outperform instruction-tuned counterparts as VLM backbones due to higher absorption rates and lower data-scaling decay; and different model families occupy distinct positions in the transfer--absorption space. The framework turns backbone selection from costly empirical sweeps into a principled, quantitative decision.

\textbf{Project Website:}
\href{https://wangq-dev.github.io/CDMScaling-Website/}{https://wangq-dev.github.io/CDMScaling-Website}
\\
\textbf{Code:}
\href{https://github.com/wangq-dev/CDMScaling}{https://github.com/wangq-dev/CDMScaling}
\\
\textbf{Model Weights:}
\href{https://modelscope.ai/profile/wangq2k}{https://modelscope.ai/profile/wangq2k}
\end{abstract}

\section{Introduction}

Scaling laws for large language models (LLMs) have proven remarkably predictive: given only compute budget, parameter count, and training data volume, one can accurately forecast not just training loss but also downstream task performance across diverse benchmarks~\citep{kaplan2020scaling, hoffmann2022training}. This predictability has transformed LLM development---practitioners can allocate resources, compare architectures, and anticipate benchmark scores before a single training run completes. 
Yet when it comes to vision-language models (VLMs), no analogous framework exists. State-of-the-art VLMs universally depend on strong LLM backbones~\citep{liu2024llavanext, bai2025qwen3, wang2025internvl3}, and practitioners widely accept that a better LLM yields a better VLM~\citep{li2024llavanext-strong}---but \emph{how much} better, on \emph{which tasks}, and \emph{for how much multimodal data}? Choosing the right LLM backbone is the single most consequential decision when building a VLM, yet it remains fundamentally unprincipled.

Compute-based scaling laws offer a natural starting point, but they fail to generalize across model families, because parameter count alone cannot capture the heterogeneous pre-training histories that differentiate modern LLMs. A more direct route is to use \emph{observable} textual benchmark performance as a capability proxy---it is publicly reproducible, requires no access to training details, and is strongly correlated with multimodal performance across diverse model families. This observation raises a central question: \textit{can we build a unified framework that directly predicts VLM performance from the textual capability of its LLM backbone?}

Addressing this question is non-trivial. LLM capability is inherently multidimensional, spanning reasoning, knowledge, instruction following, and beyond, and different dimensions may contribute heterogeneously to multimodal transfer. Furthermore, VLM performance depends jointly on LLM capability and multimodal training data volume, and their interaction must be explicitly modeled: stronger backbones may respond differently to additional multimodal data than weaker ones. Finally, LLMs from different families exhibit heterogeneous capability profiles across benchmarks, making it challenging to derive a single comparable capability measure that generalizes across families.

In this paper, we propose the \textbf{Capability-Driven Multimodal Scaling Law}, a framework that predicts multimodal model performance from the textual capability of the LLM backbone and the multimodal training data volume. To build this framework, we collect 34 LLM backbones spanning 7 model families and train over 150 VLMs under a strictly controlled multimodal training recipe, yielding a dataset of LLM--VLM pairs across diverse families and scales. 
We extract a low-dimensional capability score from more than 200 textual benchmarks to represent the basic capability of a LLM backbone, and build the multimodal scaling law by introducing transfer and absorption rates to bridge LLM capability and vision-language training.

Beyond the scaling law itself, our framework surfaces a set of actionable insights. \textbf{(i) Benchmark selection matters}: not all textual benchmarks contribute positively to multimodal transfer; some are uncorrelated or even negatively correlated with VLM performance, revealing latent benchmark-gaming behavior in certain LLMs---high textual scores that do not translate to multimodal capability. \textbf{(ii) Base models are better VLM backbones}: despite lower initial transfer rates, base LLMs exhibit higher absorption rates and lower decay, yielding superior data-scaling efficiency; instruction-tuned models incur an ``alignment tax'' on multimodal generalization. \textbf{(iii) Family-specific transfer profiles}: different model families occupy distinct positions in the (transfer, absorption) space---some families show high transfer but low absorption (strong out-of-the-box but data-inefficient), while others show the opposite, reflecting fundamental differences in pre-training strategy and benchmark saturation. The framework thus turns backbone selection from costly trial-and-error into a principled, quantitative decision.

Our main contributions are as follows:
\begin{itemize}
    \item \textbf{A capability-driven scaling law for VLMs.} We propose the Capability-Driven Multimodal Scaling Law, the first cross-family framework that predicts VLM performance from directly observable textual capability $S$ and multimodal data volume, parameterized by a per-backbone transfer rate and an absorption rate.
    \item \textbf{Insights into LLM-to-VLM transfer.} We reveal that (i) not all textual benchmarks contribute positively---some negatively correlate with multimodal performance, exposing latent benchmark gaming; (ii) base LLMs are more data-efficient VLM backbones than their instruction-tuned counterparts due to higher absorption rates and lower data-scaling decay; and (iii) model families occupy distinct positions in the transfer--absorption space, reflecting fundamental differences in pre-training strategy.
    \item \textbf{Practical applications.} The framework directly supports performance prediction for candidate backbones, optimal joint selection of backbone and data budget under a compute constraint, and hyperparameter extrapolation for efficient training configuration.
\end{itemize}

\section{Capability-Driven Scaling Laws for Multimodal Performance Prediction}
\label{sec:framework}

In this section, we establish a capability-driven framework to formulate our multimodal scaling law. Our methodology proceeds in two key modeling steps. First, we introduce a low-dimensional text capability metric $S$ to enable consistent cross-family loss fitting (Sec.~\ref{sec:loss_scaling}). Second, we derive the end-to-end vision-language performance $P$ by coupling $S$ and the multimodal data volume $D_\text{mm}$ through explicit transfer and absorption terms (Sec.~\ref{sec:mm_scaling}). Ultimately, the design of these formulations captures the core components of the underlying learning dynamics.

\subsection{Capability-Driven Scaling Law}

\subsubsection{From Classical Compute Scaling to Capability-Driven Loss Scaling}
\label{sec:loss_scaling}

The classical compute-based scaling law models the pre-training loss as a power-law function of parameter count $N$ and data volume $D$~\citep{kaplan2020scaling, hoffmann2022training}:
\begin{equation}
    L = \frac{A}{N^\alpha} + \frac{B}{D^\beta} + E,
    \label{eq:classical_scaling}
\end{equation}
where $A$, $B$, $E$, $\alpha$, and $\beta$ are fitted constants.

A natural extrapolation to the multimodal setting is to treat the LLM backbone as the model term and multimodal training data as the data term:
\begin{equation}
    L = \frac{A}{N^\alpha} + \frac{B}{D_\text{mm}^\beta} + E,
    \label{eq:mm_scaling}
\end{equation}
where $D_{\mathrm{mm}}$ denotes the amount of multimodal training data measured in billion tokens (B tokens). However, this extrapolation relies on $N$ as a proxy for backbone capability, which becomes problematic across model families. Models with identical parameter counts may differ substantially in pre-training data volume $D_\text{text}$ as well as the training strategy, and thus in actual capability. Moreover, the underlying pre-training compute $C_\text{text} \approx 6ND_\text{text}$ is rarely observable in practice, as most model providers do not disclose $D_\text{text}$.

To address the limitations of parameter count or raw pre-training compute—which fail to capture heterogeneous training histories across different model families—we propose using directly observable textual benchmarks as a capability proxy. However, because LLM capability is multidimensional, directly using raw benchmark scores as predictors leads to severe redundancy and risks overfitting on our limited VLM training pairs. To resolve this, we apply Principal Component Analysis (PCA) to extract a compact, low-dimensional representation, which can be treated as a latent capability representation optimized for predictive scaling-law modeling. 

Specifically, we construct a benchmark-model matrix $\mathbf{X} \in \mathbb{R}^{T \times M}$ containing the mean-centered scores of $M$ models across $T$ benchmarks. Prior work suggests that $\mathbf{X}$ exhibits a low-rank structure~\citep{ruan2024observational}, allowing the scores to be factorized into a $K$-dimensional capability vector $\mathbf{S}_m \in \mathbb{R}^K$ that simultaneously satisfies:

\begin{equation}
    \mathbf{S}_m \approx \boldsymbol{\theta}_f \log(C_m) +
    \boldsymbol{\nu}_f, \quad
    X_{i,m} \approx \boldsymbol{\gamma}_i^\top \mathbf{S}_m,
    \label{eq:obs_scaling}
\end{equation}
where $\boldsymbol{\theta}_f, \boldsymbol{\nu}_f \in \mathbb{R}^K$ are family-specific constants and $\boldsymbol{\gamma}_i \in \mathbb{R}^K$ are orthonormal vectors. The principal components extracted from $\mathbf{X}$ provide a natural empirical instantiation of $\mathbf{S}_m$. We therefore define the scalar capability score $S_m$ as a unit-norm linear combination of the top-$K$ principal components:
\begin{equation}
    S_m = \mathbf{w}^\top \mathbf{S}_m, \quad \|\mathbf{w}\| = 1,
    \label{eq:capability_score}
\end{equation}
where $\mathbf{w} \in \mathbb{R}^K$ is optimized jointly with the downstream scaling law. Since $S_m$ is a linear projection of $\mathbf{S}_m$, it inherits the log-linear relationship in Eq.~\ref{eq:obs_scaling}:
\begin{equation}
    S \propto \log C,
    \label{eq:s_logc}
\end{equation}

To balance the bias-variance tradeoff, we select $K$ to explain at least 95\% of the total variance in $\mathbf{X}$, while the unit-norm constraint on $\mathbf{w}$ ensures a well-defined and interpretable scale. Similar methods of latent factorization have been also employed in~\citep{ruan2024observational}, whose objective is primarily to predict unobserved textual benchmarks in a training-free manner (i.e., zero-shot benchmark prediction). In contrast, we leverage this latent representation to bridge LLM backbone and vision-language.

The capability score $S$, derived directly from benchmark performance, resolves both limitations of parameter count: it is publicly observable and, as established in Eq.~\ref{eq:s_logc}, satisfies $S \propto \log C_\text{text}$, jointly capturing the effect of both $N$ and $D_\text{text}$. Inverting gives $C_\text{text} \propto e^S$, which motivates the substitution of Eq.~\ref{eq:mm_scaling}, we arrive at the capability-driven scaling law:

\begin{equation}
    L = A \cdot e^{-\alpha S} + \frac{B}{D_\text{mm}^\beta} + E,
    \label{eq:capability_scaling}
\end{equation}

where $D_\text{mm}$ denotes the multimodal training data volume, and $S$ encodes the textual capability of the LLM backbone prior to multimodal training. We retain the power-law form $B/D_\text{mm}^\beta$ for the data term, as $D_\text{mm}$ is an explicit training hyperparameter rather than a latent capability variable, and power-law scaling with data volume is well-established empirically~\citep{kaplan2020scaling}. The residual $E$ represents the irreducible loss floor. Because $S$ is derived from benchmark performance rather than from architectural hyperparameters, Eq.~\ref{eq:capability_scaling} applies uniformly across model families without requiring family-specific recalibration.

\subsubsection{An End-to-End Predictor for Vision-Language Performance}
\label{sec:mm_scaling}

Building on the capability-driven loss scaling law above, we now introduce a direct predictor for end-to-end vision-language performance. We propose a formula that directly predicts multimodal benchmark accuracy $P$ from textual capability score $S$ and multimodal training data volume $D_\text{mm}$:

\begin{equation}
    P = \hat{A} \cdot S + \hat{B} \cdot \ln D_\text{mm} + P_0,
    \label{eq:multimodal_scaling}
\end{equation}
\begin{equation}
    \hat{B} = B_0 - B_m \cdot S,
    \label{eq:absorption_mod}
\end{equation}
where $\hat{A}$, $B_0$, $B_m$, and $P_0$ are fitted constants. 

The formulation of Eq.~\ref{eq:multimodal_scaling} is phenomenologically motivated by established empirical scaling behaviors. Specifically, downstream benchmark accuracy $P$ has been shown to scale log-linearly with training data volume ~\citep{kaplan2020scaling} and linearly with primary capability proxies~\citep{guo2025seed1}.

\subsection{Interpreting the Performance Predictor}
\label{sec:interpretation}

\subsubsection{Design Principle: Transfer and Absorption}

The two additive terms in Eq.~\ref{eq:multimodal_scaling} capture complementary sources of multimodal performance. The first term reflects the capability transferred from the textual backbone to the multimodal setting, while the second term captures the gain absorbed from multimodal training data. Under this view, multimodal performance is determined not only by the strength of the initial LLM backbone, but also by how effectively the model turns additional multimodal data into benchmark improvements.

\subsubsection{Transfer}

The first term $\hat{A} \cdot S$ reflects the \emph{transfer efficiency} from textual capability to multimodal performance: it serves as the capability-dependent starting point of multimodal training, determining the portion of multimodal performance directly attributable to the textual backbone prior to substantial multimodal adaptation. A stronger LLM backbone (higher $S$) leads to higher multimodal accuracy, with $A$ quantifying how effectively textual capability transfers to the multimodal setting.

\subsubsection{Absorption}

The second term $\hat{B} \cdot \ln D_{\text{mm}}$ in Eq.~\ref{eq:multimodal_scaling} models the performance gain from multimodal training. The logarithmic form characterizes the diminishing marginal returns of data scaling. 
Crucially, the effective data absorption rate $\hat{B}$ is modulated by the backbone's capability $S$ (Eq.~\ref{eq:absorption_mod}). We parameterize this behavior using two constants. First, $B_0$ represents the baseline absorption rate of a theoretical backbone with $S = 0$. Second, $B_m$ represents the absorption decay rate. This decay rate quantifies how quickly data-scaling efficiency diminishes as textual capability scales. 
A positive $B_m$ formalizes the empirical observation that stronger textual priors attenuate the marginal utility of additional data, which means high-capability models accelerate performance saturation and require less training data. In contrast, weaker backbones require more extensive data scaling to compensate for deficient initialization. The constant $P_0$ accounts for benchmark-specific baseline performance, such as chance-level accuracy.

\subsubsection{Transfer Tax}

While the capability score $S$ aggregates textual benchmark performance into a unified predictor, the PCA decomposition reveals that individual benchmarks contribute heterogeneously to multimodal performance. Expanding $S_m = \mathbf{w}^\top \mathbf{S}_m$ in terms of benchmark scores via the PC loading vectors $\boldsymbol{\gamma}_i$ yields an effective transfer coefficient $\lambda_i = \mathbf{w}^\top \boldsymbol{\gamma}_i$ for each benchmark $i$, such that:
\begin{equation}
    P = \hat{A} \cdot \sum_i \lambda_i X_{i,m} + \hat{B} \cdot \ln D_\text{mm} + P_0.
    \label{eq:multimodal_decomposed}
\end{equation}

We define the \textbf{transfer tax} as the set of textual capabilities for which $\hat{A} \cdot \lambda_i < 0$: improving these dimensions raises text leaderboard scores but fails to improve multimodal performance. We revisit this concept in Sec.~\ref{sec:analysis_tax} and Appendix~\ref{sec:transfer_robustness}, where we identify which benchmark dimensions exhibit positive-transfer or transfer-tax patterns, analyze their implications for LLM backbone selection, and verify the stability of these fitted associations under changes to benchmark composition and model-family mixture.

\section{Empirical Setup and Extrapolation Validation}
\label{sec:experiments}

\subsection{Models and Training Recipe}

We select a diverse set of open-source LLM backbones spanning a wide range of model families, scales, and architectures, including Qwen2.5~\citep{qwen2025qwen25technicalreport}, Qwen3~\citep{yang2025qwen3}, Llama-3.2~\citep{grattafiori2024llama}, Falcon3~\citep{Falcon3}, Gemma-2~\citep{gemmateam2024gemma2improvingopen}, DeepSeek~\citep{bi2024deepseek}, and Mistral~\citep{jiang2023mistral7b}.

All models are built upon the LLaVA-OneVision architecture~\citep{li2024llava}, comprising a SigLIP vision tower~\citep{zhai2023sigmoid}, a two-layer MLP projector, and a language backbone. A unified vision-language training recipe is applied consistently across all backbones on the Infinity-MM dataset~\citep{gu2024infinity}: Stage 1 trains only the projector ($\sim$5M samples) to align visual features with the LLM's embedding space, while Stage 2 fine-tunes the full model end-to-end ($\sim$12M samples). 
Full details on backbone selection, model list, and training hyperparameters are provided in Appendix~\ref{appendix:experimental_setup}.

\subsection{Evaluation Metrics: Textual and Multimodal Benchmarks}
\label{sec:benchmark}

\textbf{Textual Benchmarks.}
To fully build our proposed methods, we evaluate LLM backbones on more than 200 benchmarks across six dimensions: \textit{Knowledge} (world knowledge across diverse domains), \textit{Language} (reading comprehension and text understanding), \textit{Math} (quantitative reasoning and problem-solving), \textit{Reasoning} (multi-step logical and commonsense reasoning), \textit{NLI/NLU} (logical entailment and semantic understanding), and \textit{Information Extraction} (structured knowledge extraction from unstructured text, e.g., named entity recognition). All textual benchmarks are evaluated under a few-shot setting to better reflect the intrinsic capabilities of the LLM backbone.

\textbf{Multimodal Benchmarks.}
We evaluate 35 benchmarks across four dimensions: \textit{General VQA} (visual understanding and commonsense reasoning), \textit{STEM Puzzle} (vision-integrated scientific and mathematical reasoning), \textit{Document Understanding} (charts, tables, and scanned documents), and \textit{Alignment} (hallucination-free, instruction-faithful generation). All multimodal benchmarks are evaluated under a zero-shot setting. The mean score across all 35 benchmarks is reported as the \textbf{Average Multimodal Accuracy} metric.
Details of the selected benchmarks are provided in Tables~\ref{tab:text_benchmarks} and \ref{tab:multimodal_benchmarks}.

\subsection{Fitting Results: Validating the Capability-Driven Scaling Law}
\label{sec:fitting_results}
\begin{figure}[htbp]
    \centering
    \includegraphics[width=0.85\linewidth]{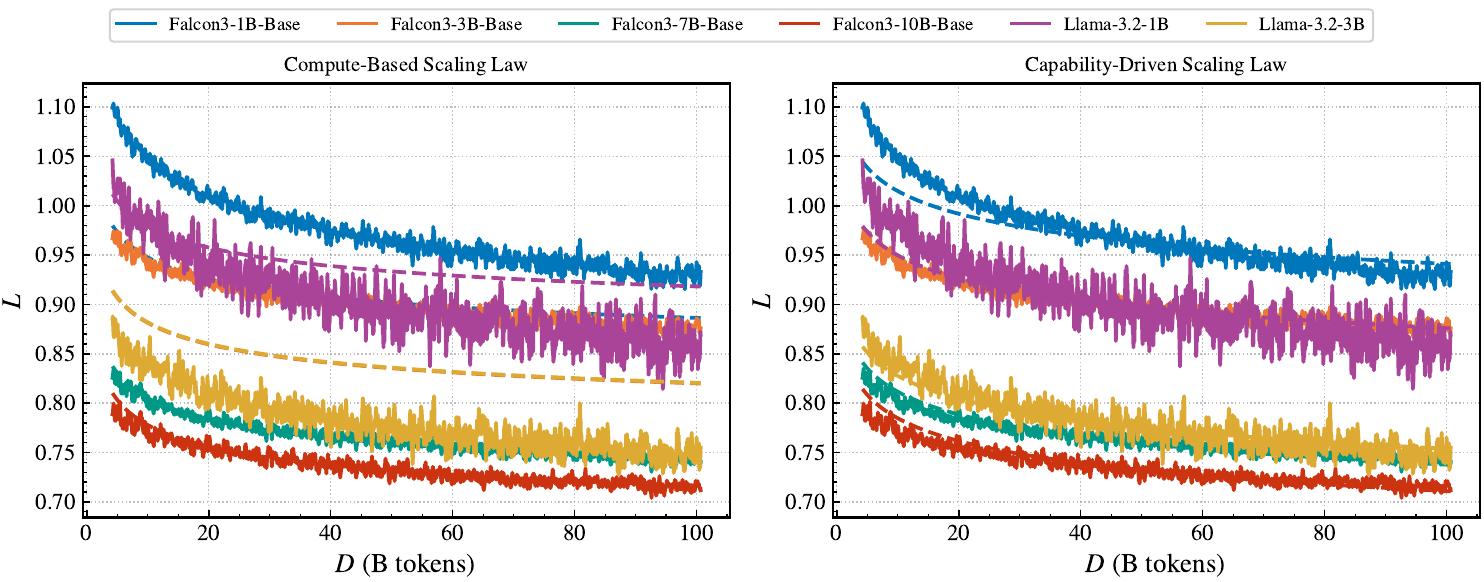}
    \caption{Comparison of two scaling law formulations fitted to the Falcon3 and Llama-3.2 base model families. 
    \textbf{Left:} Compute-based scaling law, where model scale is parameterized directly by the number of parameters $N$. 
    \textbf{Right:} Capability-driven scaling law, where the effective model scale is parameterized by principal components derived from downstream benchmark performance. Solid lines denote observed training loss trajectories and dashed lines denote fitted predictions.}
    \label{fig:loss_fitting}
\end{figure}

We compare our capability-driven scaling law (Eq.~\ref{eq:capability_scaling}) with the compute-based baseline (Eq.~\ref{eq:classical_scaling}) towards loss fitting. Specifically, we fit both laws on  Falcon3 and Llama-3.2 families. The PCA applied to the benchmark-model matrix $\mathbf{X}$ yields the first 3 principal components, which explains over 95\% of the total variance. As shown in Figure~\ref{fig:loss_fitting}, the capability-driven scaling law achieves consistent fits across both families within a single unified model, whereas the compute-based baseline requires separate fitting per family and still yields substantially higher error. Replacing $N$ with the capability score $S$ reduces the Mean Absolute Error (MAE) from 0.0383 to 0.0087 across families. 

\subsection{Predicting Multimodal Benchmark Accuracy}
\label{sec:acc_fitting}

\begin{figure*}[t]
    \centering
    \includegraphics[width=0.98\linewidth]{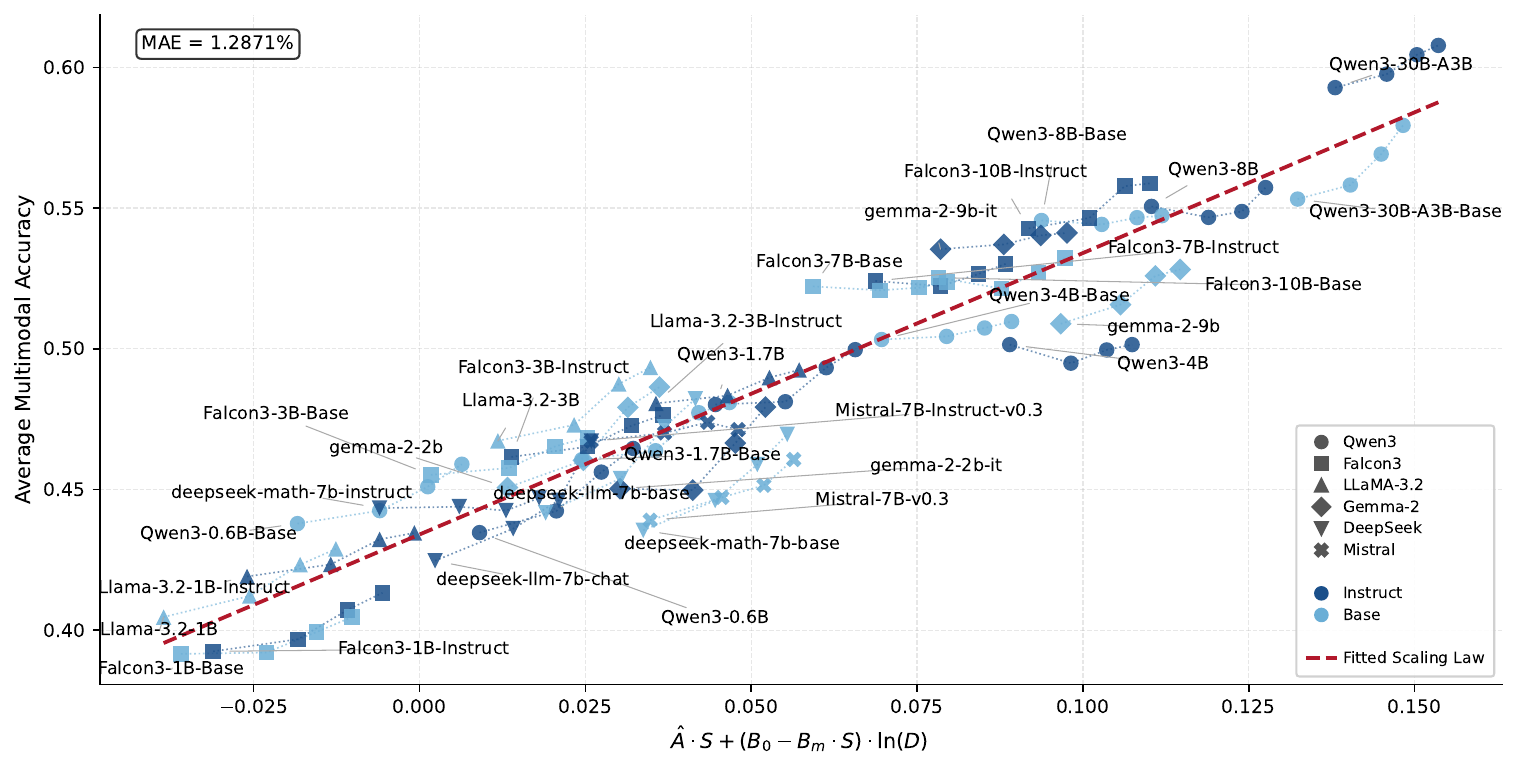}
    \caption{Multimodal accuracy scaling law fitting results across all 32 LLM backbones. For clarity, only the first 4 training checkpoints are shown for each backbone. The dashed line shows the fitted scaling law (Eq.~\ref{eq:multimodal_scaling}).}
    \label{fig:acc_fitting}
\end{figure*}

\textbf{Multimodal Accuracy Fitting.} We fit the multimodal accuracy scaling law (Eq.~\ref{eq:multimodal_scaling}) across all 32 LLM backbones, following the fitting procedure detailed in Appendix~\ref{sec:appendix_algorithm}. As shown in Figure~\ref{fig:acc_fitting}, the predicted accuracy trajectories closely follow the observed trends across all model families and scales, with the fitted scaling law achieving a MAE of 1.2870\% without requiring family-specific recalibration. We further analyze the validity and robustness of the learned capability representation. Appendix~\ref{sec:alternative_predictors} compares the proposed representation with simpler predictors, while Appendix~\ref{sec:correlation_robustness} evaluates its robustness to highly correlated textual benchmarks.

\begin{figure}[ht]
    \centering
    \begin{subfigure}[t]{0.48\linewidth}
        \centering
        \includegraphics[width=\linewidth]{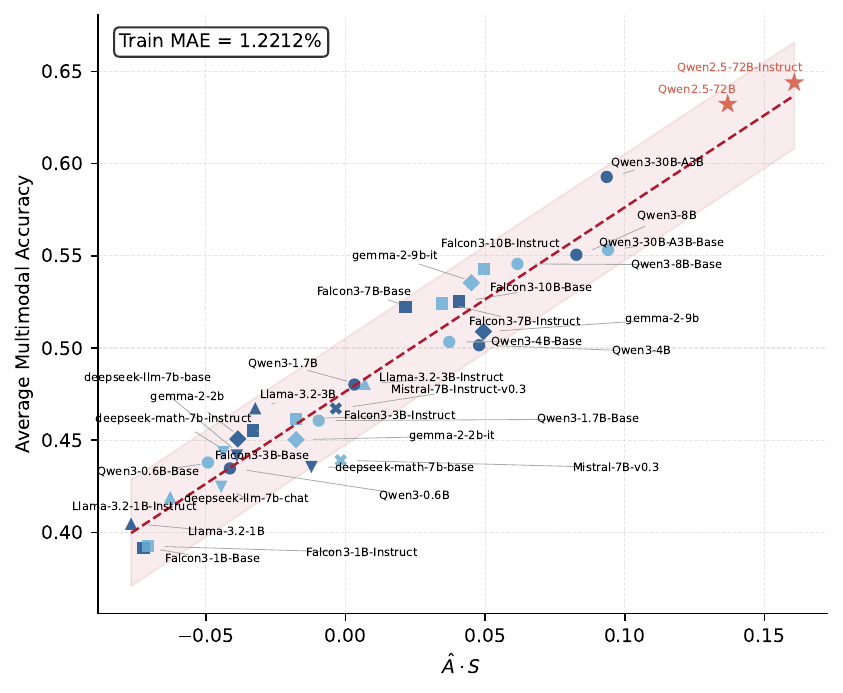}
        \caption{Predicted vs.\ observed multimodal accuracy at the starting point ($D_{mm} \approx 12.5$B tokens) for the Base and Instruct variants of Qwen2.5‑72B. The dashed line denotes the fitted scaling law, and the shaded region shows a prediction error band ($\pm1.96\sigma$).}
        \label{fig:extrap_A}
    \end{subfigure}
    \hfill
    \begin{subfigure}[t]{0.48\linewidth}
        \centering
        \includegraphics[width=\linewidth]{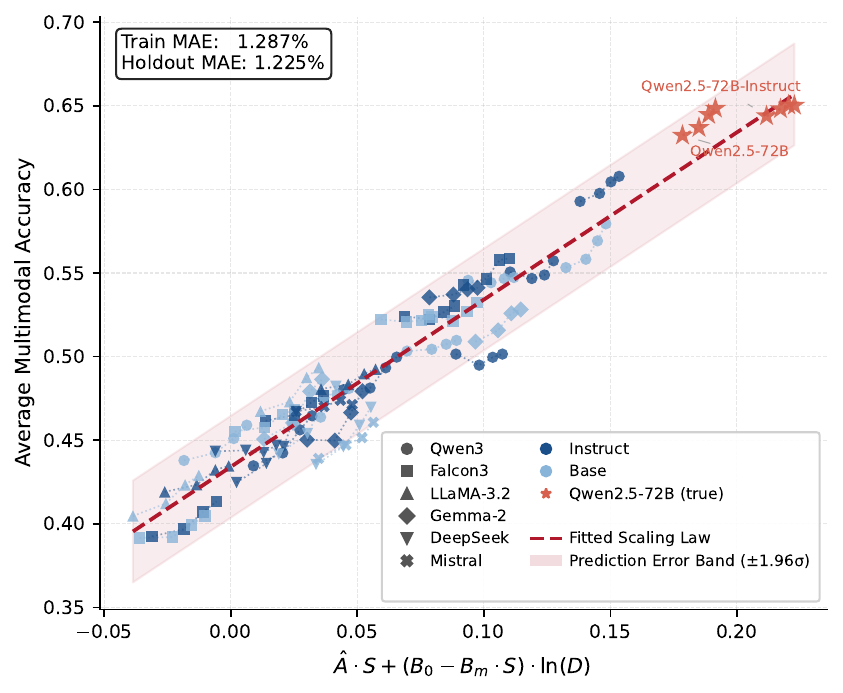}
        \caption{Predicted vs.\ observed multimodal accuracy trajectories for multiple model families, including Qwen2.5‑72B holdout variants. The dashed line denotes the fitted scaling law, and the shaded region shows a prediction error band ($\pm1.96\sigma$).}
        \label{fig:extrap_B}
    \end{subfigure}
    \caption{Comparison of predicted vs.\ observed multimodal accuracy under different evaluation settings, with fitted scaling laws and prediction error bands ($\pm 1.96\sigma$), where $\sigma$ is the standard deviation of the training residuals.}
    \label{fig:extrap_combined}
\end{figure}

\textbf{Validating Transfer.} To evaluate whether the fitted transfer coefficient $\hat{A}$ generalizes to unseen backbones at larger scales, we fix the multimodal training data $D_{mm}$ and directly fit $\hat{P} = \hat{A} S + P_0'$ for validation.
We use the fitted law to predict the multimodal accuracy of Qwen2.5-72B, whose parameter count far exceeds that of any model seen during fitting. As shown in Figure~\ref{fig:extrap_A}, the predicted accuracy closely matches the observed values, with an absolute error of 1.92\% for the Base variant and 0.70\% for the Instruct variant, demonstrating that $\hat{A}$ captures transfer efficiency reliably even for model scales beyond the training distribution.

\textbf{Validating Absorption.} To validate the rationality of the absorption coefficient $\hat{B}$, we first predict the full accuracy trajectorys of Qwen2.5-72B across all training checkpoints using the complete prediction formula (Eq.~\ref{eq:multimodal_scaling}). 
As shown in Figure~\ref{fig:extrap_B}, the predicted region correctly covers the target models, achieving a combined holdout MAE of only 1.225\% on both Base and Instruct variants of Qwen2.5-72B. 

In addition, we further analyze the ablation of adding the decay rate ($B_m \cdot S$), which is shown in Table~\ref{tab:absorption_ablation}. The results show that for models with moderate capability scores ($S \in [-0.369, 0.339]$), the two variants perform comparably since the score $S$ is close to zero. However, when it comes to predicting models with substantially higher capability scores ($S > 0.6$), adding the decay rate can effectively reduce the overall MAE (reducing the MAE of Qwen2.5-72B-Base from 2.374\% to 2.055\%). These results collectively demonstrate the importance of the interaction term $B_m \cdot S$ as $S$ grows larger: stronger backbones exhibit systematically lower data absorption rates, and failing to account for this modulation leads to progressively larger prediction errors when extrapolating beyond the training distribution. This validates the necessity of the capability-modulated absorption term in Eq.~\ref{eq:multimodal_scaling} for accurate out-of-distribution trajectory prediction.

\begin{table}[h]
\centering
\caption{Comparison of in-distribution ($\le$8B) and out-of-distribution (72B) prediction errors (MAE \%) with and without capability-modulated absorption ($B_m S$).}
\label{tab:absorption_ablation}
\resizebox{0.75\linewidth}{!}{%
\begin{tabular}{lccc}
\toprule
\textbf{Variant} & \textbf{$\leq$8B MAE $\downarrow$} & \textbf{72B-Base MAE $\downarrow$} & \textbf{72B-Inst MAE $\downarrow$} \\
 & $S \in [-0.369, 0.339]$ & \multicolumn{2}{c}{$S > 0.6$} \\
\midrule
$B_0 \cdot \ln D_\text{mm}$  & 1.266\% & 2.374\% & 0.476\% \\
$(B_0 - B_m \cdot S) \cdot \ln D_\text{mm}$ & 1.259\% & 2.055\% & 0.394\% \\
\bottomrule
\end{tabular}%
}
\end{table}

\subsection{Cross-Family Generalization}
\label{sec:cross_family}
\begin{figure*}[t]
    \centering
    \includegraphics[width=0.98\linewidth]{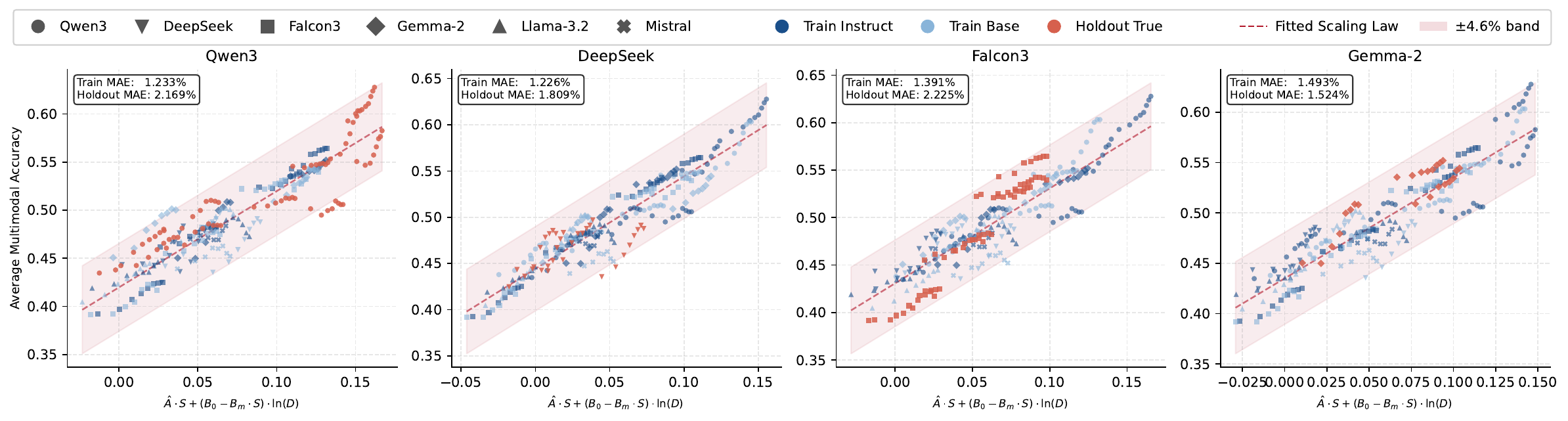}
    \caption{Cross-family generalization using the capability-driven formula (Eq. ~\ref{eq:multimodal_scaling}). }
    \label{fig:cross_family_comparison}
\end{figure*}

To validate the robustness of our methods across different model families, we systematically evaluate its cross-family generalization performance using leave-one-family-out validation. Specifically, we sequentially select each of the four representative model families (Qwen3, DeepSeek, Falcon3, and Gemma-2) as a held-out target, and fit our capability-driven scaling law on the remaining in-domain families to predict the entire trajectories of the held-out family. As illustrated in Figure~\ref{fig:cross_family_comparison}, our method consistently yields highly accurate extrapolation trajectories across all target lineages, with the holdout checkpoints closely following the predicted curves. Although a small number of checkpoints exhibit noticeable deviations from the predicted curves, the overall prediction errors remain within an acceptable range, preserving the global scaling trend and relative performance ordering across model scales.
This consistent generalizability demonstrates that our strategy successfully bypasses heterogeneous architectural and pre-training recipe differences, establishing a robust cross-family framework for multimodal performance prediction.

\section{Understanding LLM-to-VLM Transfer}

In this section, we analyze the transfer dynamics from LLM to VLM through four complementary lenses: the identification of beneficial and harmful textual capabilities (Sec.~\ref{sec:analysis_tax}), the paradoxical behavior of instruction-tuned backbones (Sec.~\ref{sec:analysis_it}), a cross-family comparison of transfer dynamics via the fitted scaling coefficients (Sec.~\ref{sec:analysis_family}), and the extrapolation of optimal training hyperparameters across model scales (Sec.~\ref{sec:hyperparam_extrapolation}). We further investigate whether these transfer and absorption patterns differ across multimodal capability dimensions by fitting the scaling law separately on the benchmark categories summarized in Table~\ref{tab:multimodal_benchmarks} (Appendix~\ref{sec:category_scaling}).

\subsection{Analyzing LLM-to-VLM Transfer Efficiency}
\label{sec:analysis_tax} 
The transfer coefficient $\lambda_j = \mathbf{w}^\top \boldsymbol{\gamma}_j$ in Eq.~\ref{eq:multimodal_decomposed} measures how the $j$-th textual benchmark contributes to downstream multimodal performance. Specifically, positive coefficients ($\lambda_j > 0$) identify textual benchmark dimensions that are positively associated with multimodal transfer, such as structured reasoning (\textit{dyck\_languages\_hard}, \textit{matrixshapes}) and factual knowledge (\textit{mmlu\_stem}). Neutral coefficients ($\lambda_j \approx 0$) stem from dimensional orthogonality (e.g., purely syntactic \textit{tense} detection) or saturated performance (e.g., \textit{piqa}). Crucially, negative coefficients ($\lambda_j < 0$) represent a transfer-tax pattern: these benchmark dimensions exhibit negative fitted associations with multimodal transfer. This suggests that improvements on certain text-specific benchmarks may not consistently translate into multimodal gains. One possible explanation is that optimization toward specific textual evaluation formats may be less aligned with multimodal requirements. Selecting backbones by unweighted leaderboard averages can thus be misleading, as scores can be artificially inflated by gaming negative-transfer benchmarks; practitioners should instead prioritize capabilities with significantly positive coefficients ($\lambda_j \gg 0$). The full list of the fitted transfer coefficients for all textual benchmarks is provided in Table~\ref{tab:complete_coefficients}.

\subsection{The Instruction-Tuning Disparity in Multimodal Training}
\label{sec:analysis_it}

Instruction-tuned (IT/Chat) language models frequently yield suboptimal vision-language model (VLM) backbones compared to their base counterparts, despite their superior textual benchmark performance. To systematically analyze this instruction-tuning disparity, we summarize the fitted scaling parameters of Eq.~\ref{eq:multimodal_scaling} for all the Base and Instruct backbones separately, as shown in Table~\ref{tab:base_vs_chat}.

The fitted parameters characterize the baseline divergence in scaling behavior between the two regimes. Although instruction tuning does not attenuate the initial transfer efficiency—as evidenced by instruct models' higher transfer slope ($A = 0.254$ vs. $0.212$)—it imposes a significant constraint on subsequent data scaling. Specifically, the data absorption decay rate $B_m$ of Instruct models is $1.33\times$ higher than that of Base models ($0.0104$ vs. $0.0078$). This accelerated decay indicates that as textual capability $S$ increases, Instruct models' capacity to absorb multimodal data saturates faster than that of Base models. We attribute this phenomenon to the alignment tax: over-optimizing the representational space for text-specific instruction formatting (such as prior suppression, Sec.~\ref{sec:analysis_tax}) constrains the latent space's geometric flexibility, reducing its residual capacity to align with continuous visual embeddings. Consequently, the fitted aggregate scaling law indicates that Base backbones exhibit stronger scaling advantages in large-data regimes (high $D_{mm}$) and at higher backbone capacities, where their higher absorption rate ($\bar{B} = 0.0161$ vs. $0.0147$) overcomes Instruct models' initial transfer head start. Nevertheless, this aggregate trend does not imply universal superiority: Instruct backbones remain competitive in low-resource regimes (low $D_{mm}$) and in specific model families with more favorable scaling behavior. We further investigate these regime-dependent behaviors through a pair-wise crossover analysis in Appendix~\ref{sec:crossover_analysis}.

\begin{table}[h]
\centering
\caption{Comparison of scaling and transfer parameters between Base and Instruct backbones fitted via Eq.~\ref{eq:multimodal_scaling}.}
\label{tab:base_vs_chat}
\resizebox{0.6\linewidth}{!}{%
\begin{tabular}{lccccc}
\toprule
\textbf{Type} & $A$ & $B_0$ ($10^{-2}$) & $B_m$ ($10^{-2}$) & $\bar{B}$ ($10^{-2}$) & MAE (\%) \\ 
\midrule
Base & 0.212 & 1.60 & 0.78 & 1.61 & 1.33 \\
Chat & 0.254 & 1.49 & 1.04 & 1.47 & 1.11 \\ 
\bottomrule
\end{tabular}%
}
\end{table}

\subsection{Cross-Family Analysis: Heterogeneity in Transfer and Absorption}
\label{sec:analysis_family}

While Sec.~\ref{sec:analysis_it} establishes systemic Base-Instruct differences, architectural variations, pretraining corpora, and training recipes also dictate distinct transfer dynamics. We resolve these intra-family behaviors by separately fitting for four representative model families with sufficient sample sizes, as summarized in Table \ref{tab:family_fitting}.

Interestingly, the results reveal distinct trade-offs between text prior utilization and multimodal scaling potential across different model families. Specifically, \textbf{Llama} exhibits a ``high-transfer, low-capacity'' regime, yielding the highest transfer slope ($A = 0.361$) but suffering from a low basic absorption ($B_0 = 0.0028$) and high decay rate ($B_m = 0.0455$), which leads to a rapid scaling saturation. On the contrary, \textbf{Qwen3} represents a ``low-transfer, high-capacity'' regime, showing a low transfer slope ($A = 0.246$) while performing well in absorption ($\bar{B} = 0.0126$). The relatively low $A$ likely stems from over-optimizations on textual benchmarks, and these optimizations are difficult to be fully transferred into multimodal training. However, because of the strong model capacity and representation, Qwen3 family shows highly robust towards extensive data scaling.

\begin{table}[h]
\centering
\caption{Fitted scaling and transfer parameters of Eq.~\ref{eq:multimodal_scaling} across representative model families. For consistency, $B_0, B_m$, and $\bar{B}$ are scaled by $10^2$.}
\label{tab:family_fitting}
\resizebox{0.65\linewidth}{!}{%
\begin{tabular}{lccccc}
\toprule
\textbf{Family} & $A$ & $B_0$ ($10^{-2}$) & $B_m$ ($10^{-2}$) & $\bar{B}$ ($10^{-2}$) & MAE (\%) \\ 
\midrule
Llama   & 0.361 & 0.28 & 4.55 & 1.22 & 0.90 \\
Gemma   & 0.324 & 2.46 & 4.51 & 2.13 & 1.02 \\
Falcon3 & 0.288 & 1.12 & 1.33 & 1.17 & 0.60 \\
Qwen3   & 0.246 & 1.40 & 2.19 & 1.26 & 1.11 \\ 
\bottomrule
\end{tabular}%
}
\end{table}

\subsection{Extrapolation of Hyperparameters Across Scales}
\label{sec:hyperparam_extrapolation}

\begin{figure}[!htp]
    \centering
    \includegraphics[width=0.6\linewidth]{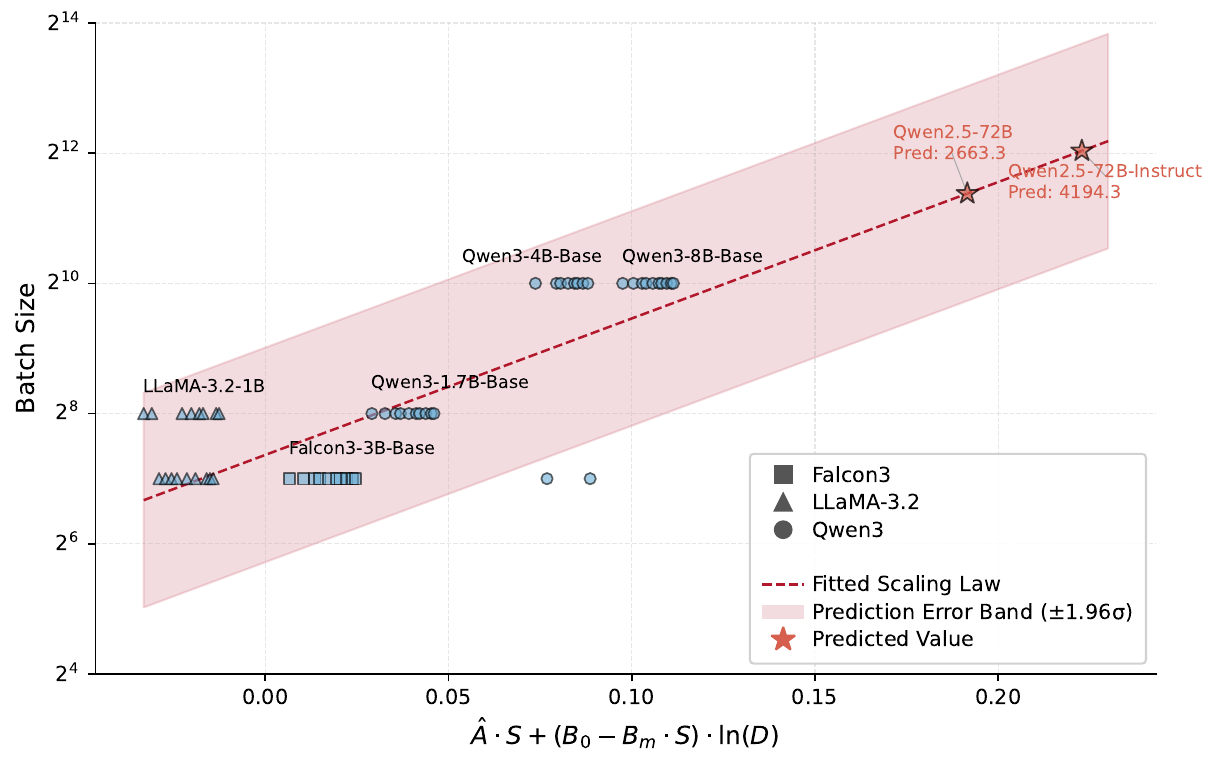}
    \caption{Relationship between the optimal hyperparameter batch size and the proposed composite scaling metric in multimodal finetuning. The predicted optimal batch sizes for Qwen2.5-72B and Qwen2.5-72B-Instruct at 50B tokens are annotated in the figure.}
    \label{fig:batch_size_extrapolation}
\end{figure}

Beyond accuracy prediction, the Capability-Driven Multimodal Scaling Law also provides a reliable foundation for inferring optimal training hyperparameters across different model scales. To obtain the guidance, controlled finetuning experiments were conducted on five representative multimodal backbones: Llama-3.2-1B, Falcon3-3B-Base, Qwen3-1.7B-Base, Qwen3-4B-Base, and Qwen3-8B-Base. For each backbone, training was performed under five batch size configurations in \{128, 256, 512, 1024, 2048\}. The experiments started at approximately 16B tokens, up to around 50B, ensuring consistent compute budgets across settings. Along the horizontal axis, we compared the loss values across the five batch size configurations at each identical metric value. The batch size yielding the lowest loss was plotted as the point’s vertical coordinate in the figure.

As shown in Figure~\ref{fig:batch_size_extrapolation}, we find that the optimal batch size for multimodal finetuning increases with the composite scaling metric $\hat{A} \cdot S + (B_0 - B_m \cdot S) \cdot \ln D_\text{mm}$, indicating that models with stronger textual capability and larger multimodal dataset size can process bigger data batches more efficiently. Leveraging this composite metric allows us to estimate suitable hyperparameters for unseen backbones directly from their benchmark scores and training data specifications, without costly tuning runs.

Compared with the compute-based scaling law, which uses parameter count $N$ as the predictor, the composite metric avoids misleading ordering of hyperparameter optima. For example, compute-based scaling would predict that Falcon3-3B-Base requires a larger optimal batch size than Qwen3-1.7B-Base due to its higher parameter count. In practice, the opposite holds: Qwen3-1.7B-Base, with a higher composite metric value, benefits from a larger batch size. This underscores that the proposed composite scaling metric captures backbone capability and data scaling effects more faithfully, enabling correct hyperparameter extrapolation across model families and scales.

\section{Conclusion}

In this paper, we proposed the Capability-Driven Multimodal Scaling Law, a predictable framework that shifts VLM performance forecasting from parameter-driven approximations to directly observable textual capabilities. Backed by an extensive empirical training suite totaling 25,000 H800 GPU-days across 34 LLM backbones, our law successfully predicts downstream multimodal trajectories up to 72B-scale models and entirely held-out lineages. Beyond prediction, our systematic analysis has exposed a "transfer tax" on gamable benchmarks, characterized the data-scaling advantages of base backbones over instruction-tuned variants due to lower absorption decay, and mapped model families onto a distinct transfer--absorption trade-off space. Ultimately, this work turns costly backbone selection sweeps into a principled, zero-shot quantitative decision, providing a scalable blueprint for future multi-modal scaling research.

\section*{Limitations}

While the capability-driven scaling law establishes a predictable and unified framework for VLM performance, we position our controlled empirical setup as a pioneering first step, focusing on the primary bottlenecks of multimodal scaling while keeping auxiliary variables regulated. Two strategic boundaries of our current framework warrant discussion:

First, all experiments were systematically conducted under a unified, late-fusion training recipe (based on LLaVA-OneVision) to cleanly isolate the mathematical contribution of the LLM backbone. Although this standardized pipeline serves as a rigorous baseline for our cross-family study, we observe strong cross-family generalization across diverse LLM backbone architectures. A natural and exciting avenue for future work is to extend and validate this framework on alternative fusion paradigms, particularly emerging \textbf{early-fusion} or native mixed-modal architectures, where visual and textual tokens are integrated from the very first layer.

Second, our framework treats the pre-trained vision encoder as a fixed component. In the context of modern VLMs, the frozen visual encoder effectively functions as a static \textbf{visual vocabulary}. Empirically, the LLM backbone serves as the core reasoning engine, accounting for the vast majority of computational cost and behavioral complexity, whereas scaling the visual encoder often yields highly predictable and localized saturation. To establish a robust pioneering law, it was methodologically vital to first crack the most challenging variable—the LLM backbone capability. Incorporating the co-scaling dynamics of this "visual vocabulary" (e.g., scaling encoder capacity or input resolutions) represents a straightforward and structured extension of our current law.

\bibliographystyle{unsrtnat}
\bibliography{ref}


\appendix
\section{Preliminary Observations}

\subsection{Correlation Between Textual and Multimodal Capabilities}

Intuitively, a stronger LLM backbone should yield a more capable VLM. To empirically validate this, we collect and analyze the evaluation results of 17 representative VLMs and their corresponding LLMs from the OpenCompass leaderboard.

\begin{figure}
    \centering
    \includegraphics[width=0.7\linewidth]{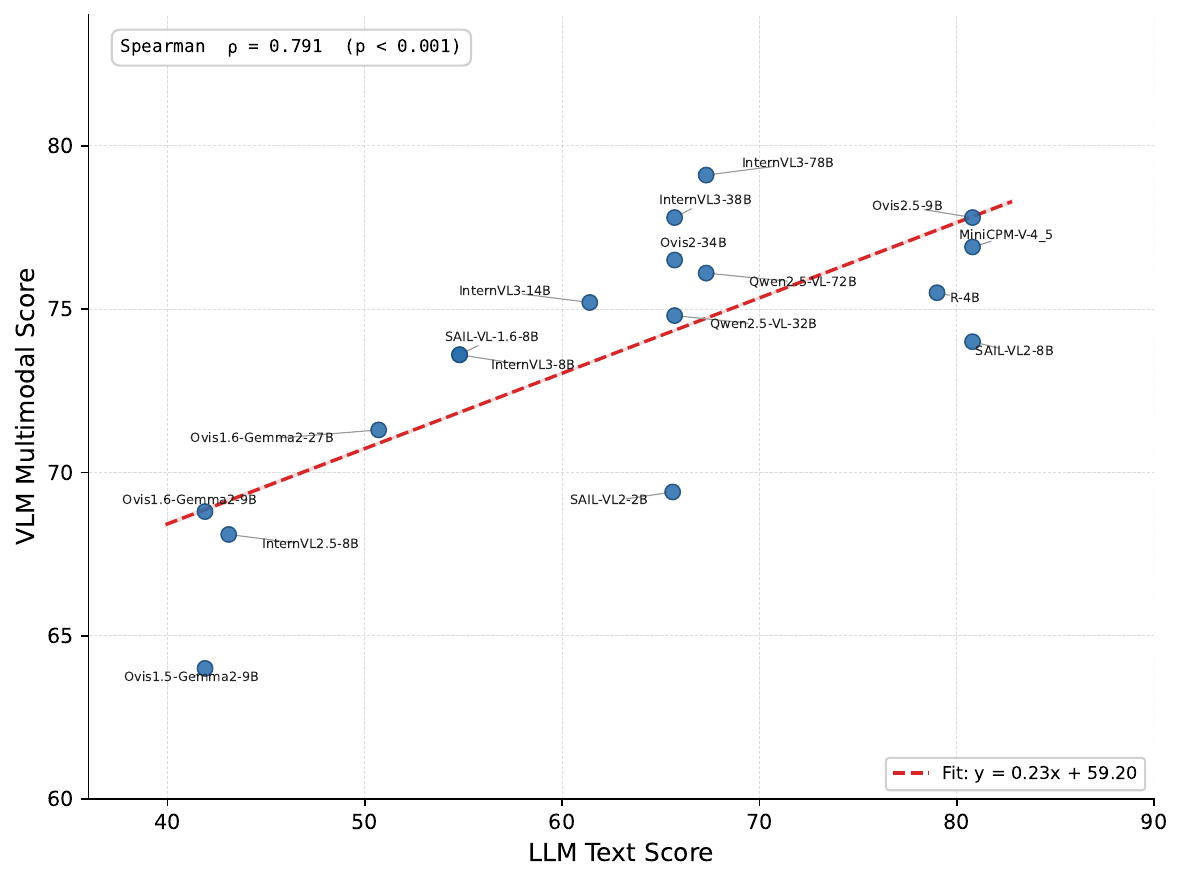}
    \caption{Correlation between LLM text scores and VLM multimodal scores across 17 models. Each point represents an individual model. The dashed line indicates the linear regression fit. Spearman's rank correlation coefficient ($\rho = 0.791$, $p < 0.001$) is shown in the upper left, and the fitted equation ($y = 0.23x + 59.20$) is located in the lower right. All evaluation scores are sourced from the OpenCompass leaderboard \citep{2023opencompass}, available at \url{https://opencompass.org.cn}.}
    \label{fig:preliminary_observations}
\end{figure}

Figure~\ref{fig:preliminary_observations} illustrates the correlation between models' textual and multimodal proficiencies. Specifically, the textual capability (x-axis) is computed as the average score across eight benchmarks spanning complementary capability dimensions: MMLU-Pro~\citep{wang2024mmlu}, GPQA-Diamond~\citep{rein2023gpqa}, BBH~\citep{suzgun2022challenging}, MATH-500~\citep{hendrycksmath2021}, AIME~\citep{dekoninck2026matharena}, LiveCodeBench~\citep{jain2025livecodebench}, HumanEval~\citep{chen2021codex}, and IFEval~\citep{zhou2023instructionfollowingevaluationlargelanguage}. The multimodal performance (y-axis) is calculated as the average score across eight prominent benchmarks: MMBench v1.1 Chinese and English test sets~\citep{liu2024mmbench}, MMStar~\citep{chen2024we}, MMMU val set~\citep{yue2023mmmu}, MathVista test-mini split~\citep{lu2024mathvista}, HallusionBench~\citep{Guan_2024_CVPR}, AI2D test set~\citep{kembhavi2016diagram}, OCRBench~\citep{Liu_2024}, and MMVet~\citep{yu2024mm}. We observe a strong positive correlation between the foundational textual capabilities of the LLMs and the performance of their VLM counterparts, suggesting that a stronger textual foundation generally leads to better multimodal performance.

\subsection{The Limitation of Compute-Based Fitting}

\begin{figure}[h]
    \centering
    \begin{subfigure}[t]{0.48\linewidth}
        \centering
        \includegraphics[width=\linewidth]{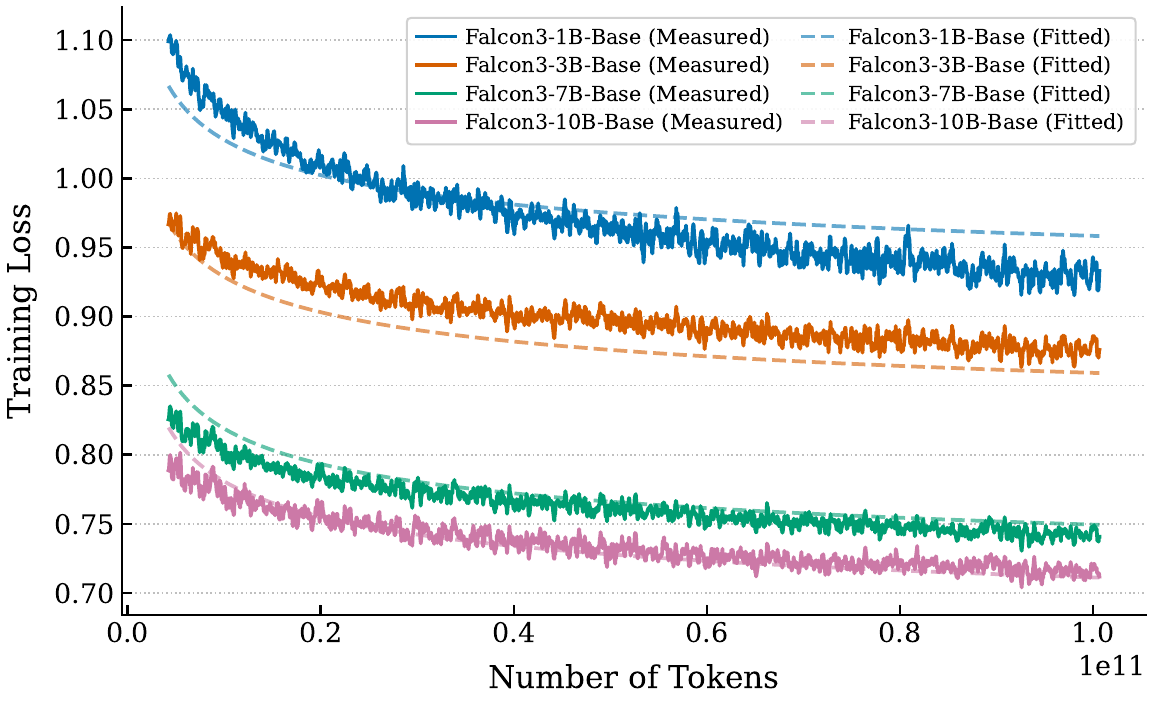}
        \caption{Falcon3 family scaling law fitting.}
        \label{fig:loss_of_falcon3}
    \end{subfigure}
    \hfill
    \begin{subfigure}[t]{0.48\linewidth}
        \centering
        \includegraphics[width=\linewidth]{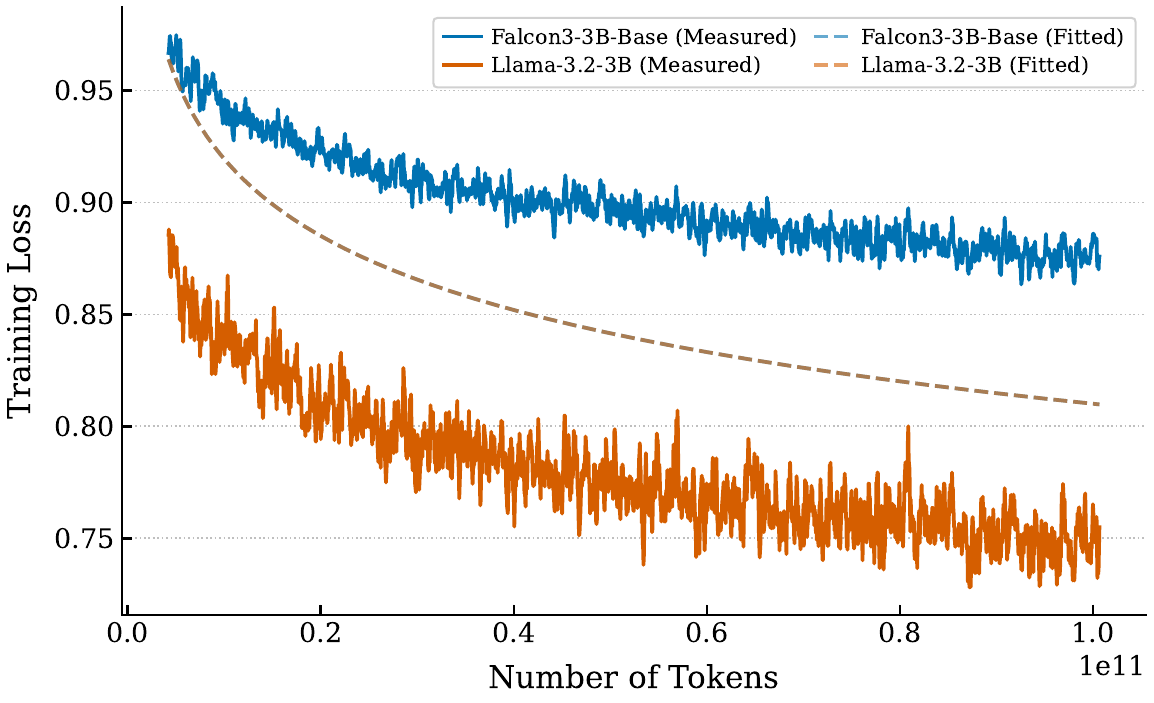}
        \caption{Cross-family comparison: Llama-3.2-3B vs.\ Falcon3-3B-Base.}
        \label{fig:loss_of_falcon3_and_llama3dot2}
    \end{subfigure}
    \caption{
        Multimodal training loss scaling law fitting and cross-family comparison, using the methodology of~\citet{hoffmann2022training}. 
        (a) Falcon3 family fitted with a compute-based scaling law. 
        (b) Despite similar parameter counts, Llama-3.2-3B and Falcon3-3B-Base show divergent loss trajectories that cannot be captured by a single compute-based scaling curve.
    }
    \label{fig:loss_combined}
\end{figure}

A natural hypothesis is that multimodal training loss follows a scaling law governed by model size and training data volume, consistent with well-validated scaling laws in language modeling~\citep{hoffmann2022training, kaplan2020scaling}. To verify this hypothesis, we train a series of VLMs across multiple model families using a unified training recipe (Appendix~\ref{training_details}), and model the multimodal training loss as a power-law function of parameter count $N$ and multimodal training token volume $D_\text{mm}$, following the fitting methodology of~\citet{hoffmann2022training}:

\begin{equation}
    L(N, D) = \frac{A}{N^{\alpha}} + \frac{B}{D_\text{mm}^{\beta}} + E
    \label{eq:scaling_law}
\end{equation}
\noindent where $A$, $B$, $\alpha$, $\beta$, and $E$ are fitted constants, and $E$ represents the irreducible loss. As shown in Figure~\ref{fig:loss_of_falcon3}, this formulation fits well within the same model family: scaling up from Falcon3-1B-Base to Falcon3-10B-Base~\citep{Falcon3} yields a steady reduction in multimodal training loss consistent with Eq.~\ref{eq:scaling_law}. The fitting procedure is detailed in Appendix~\ref{appendix:method_for_fitting_loss}.

However, this regularity breaks down across model families. The root cause is that $N$ alone is an incomplete proxy for LLM capability: it ignores the pre-training data volume $D_\text{text}$, which jointly determines total training compute $C_\text{text} \approx 6ND_\text{text}$~\citep{kaplan2020scaling}. Two models with identical $N$ but different $D_\text{text}$ therefore operate at different effective compute scales, resulting in substantially different textual capabilities that Eq.~\ref{eq:scaling_law} is structurally blind to.

This is concretely illustrated by Llama-3.2-3B~\citep{grattafiori2024llama} and Falcon3-3B-Base~\citep{Falcon3}: despite nearly identical parameter counts, their divergent pre-training histories produce substantially different textual capabilities, leading to divergent VLM loss trajectories that a unified $N$-$D$ scaling law fails to fit (Figure~\ref{fig:loss_of_falcon3_and_llama3dot2}).

This drives us to shift from parameter-driven to capability-driven fitting: rather than using $N$ as a surrogate for model quality, we directly measure the observable textual capability of an LLM as a unified proxy that implicitly encodes the joint effect of $N$, $D_\text{text}$, and training quality. The formal definition is presented in Sec.~\ref{sec:framework}.

\section{Related Works}

\textbf{Vision Language Models.} Driven by rapid advancements in LLMs \citep{grattafiori2024llama, yang2025qwen3, team2025gemma}, VLMs \citep{li2026qwen3, wang2025internvl3, wu2024deepseek, guo2025seed1} have achieved remarkable progress in recent years. Built upon pretrained LLMs, mainstream VLM architectures such as Qwen3-VL \citep{li2026qwen3} and Seed1.5-VL \citep{guo2025seed1} integrate visual tokens by projecting them into the language model's embedding space as sequential inputs, following paradigms established by LLaVA \citep{liu2023visual} and MiniGPT-4 \citep{zhu2023minigpt}. However, despite this architectural dependency, the impact of the underlying LLM's intrinsic capabilities on VLM performance remains largely unexplored. While extensive research has focused on optimizing vision encoders and cross-modal alignment mechanisms such as refining CLIP-based encoders \citep{tong2024eyes}, designing Q-former connectors \citep{dai2023instructblip}, and curating instruction-tuning data \citep{wei2023instructiongpt, gu2024infinity}, the systematic investigation into how fundamental LLM attributes propagate to and shape multimodal behaviors is still lacking.

\textbf{Scaling Laws.} Scaling laws establish predictive frameworks for allocating computational resources to maximize model performance. \citet{kaplan2020scaling} demonstrate power-law scaling in autoregressive language models, where loss improves predictably with model size, data, and compute. However, \citet{hoffmann2022training} challenge this through refined training configurations, arguing that model size and training tokens must be scaled equally---contradicting Kaplan's earlier conclusions. Downstream performance has also been examined: \citet{gadre2024language} study over-training effects on task-specific metrics, while others investigate scaling laws for translation and agentic benchmarks \citep{isik2024scaling, ruan2024observational}. Beyond training, inference-time scaling \citep{sardana2023beyond} reveals critical trade-offs: for high inference demand, models should be over-trained beyond Chinchilla-optimal points using smaller parameters but more tokens per parameter to minimize total deployment costs. Under data constraints, moderate repetition has minimal impact, but excessive repetition causes multi-epoch degradation and diminishing returns \citep{muennighoff2023scaling, xue2023repeat}. However, most scaling laws are primarily established for large language models, with multimodal counterparts remaining underexplored. Notably, \citet{aghajanyan2023scaling} examined multimodal models that tokenize text, speech, and image modalities into discrete tokens for unified generation. In parallel, recent work on native multimodal models has focused on early-fusion architectures \citep{shukor2025scaling}. In contrast, we systematically investigate late-fusion models that process image-text inputs and generate text outputs, focusing on the scaling dynamics of LLM backbones while keeping vision encoders frozen.

\section{Experimental Setup}
\label{appendix:experimental_setup}

To validate the proposed capability extrapolation framework from LLMs to VLMs, and to thoroughly investigate the behavior of different model families, parameter scales, and alignment stages during multimodal transfer, we design a strictly controlled experimental framework. The core objective of this study is to isolate and quantify the independent impacts of intrinsic LLM capabilities and the volume of multimodal training data on the final VLM performance. Therefore, across all experiments, we maintain strict consistency in the multimodal alignment architecture and the training recipe, treating only the selection of the LLM and the multimodal training data volume as independent variables.

\subsection{Model Selection}

To systematically evaluate how capabilities transfer from LLMs to VLMs and to validate our proposed extrapolation framework, we carefully curate a diverse set of base models. Rather than relying on a homogenous group of models, our selection is designed to control specific variables and is driven by three primary considerations:

\textbf{Cross-Family Comparison and Decoupling Compute Scaling} Compute scaling laws often fail across different model families. To decouple performance from parameter counts, we pair models with identical sizes but distinct text capabilities (e.g., Llama-3.2-1B vs. Falcon3-1B). This verifies our hypothesis: intrinsic text capability dictates the multimodal starting point.

\textbf{Alignment Stages Exploration} To analyze how different text alignment stages affect multimodal transfer, we systematically select paired pre-trained and instruction-tuned versions of the same underlying LLMs. This paired setup allows us to quantitatively compare their transfer behaviors and investigate how the specific text capabilities altered during the instruction-tuning stage influence subsequent visual-language alignment.

\textbf{Scale Span} Our selected models span parameter sizes from 0.6B to 72B and include Mixture-of-Experts (MoE) architectures. 

Detailed specifications and family affiliations of all selected models are summarized in Table~\ref{tab:model_list}.

\begin{table*}[t]
    \centering
    \caption{Summary of all 34 LLM backbones used in this study. Model names correspond to their Hugging Face identifiers.}
    \label{tab:model_list}
    \begin{tabular}{lllc}
        \toprule
        \textbf{Family} & \textbf{Model} & \textbf{Type} & \textbf{Size} \\
        \midrule
        \multirow{10}{*}{Qwen3}
            & Qwen3-0.6B-Base        & Base & 0.6B      \\
            & Qwen3-0.6B             & Instruct & 0.6B      \\
            & Qwen3-1.7B-Base        & Base & 1.7B      \\
            & Qwen3-1.7B             & Instruct & 1.7B      \\
            & Qwen3-4B-Base          & Base & 4B        \\
            & Qwen3-4B               & Instruct & 4B        \\
            & Qwen3-8B-Base          & Base & 8B        \\
            & Qwen3-8B               & Instruct & 8B        \\
            & Qwen3-30B-A3B-Base     & Base & 30B (A3B) \\
            & Qwen3-30B-A3B          & Instruct & 30B (A3B) \\
        \midrule
        \multirow{2}{*}{Qwen2.5}
            & Qwen2.5-72B            & Base & 72B \\
            & Qwen2.5-72B-Instruct   & Instruct & 72B \\
        \midrule
        \multirow{8}{*}{Falcon3}
            & Falcon3-1B-Base        & Base & 1B  \\
            & Falcon3-1B-Instruct    & Instruct & 1B  \\
            & Falcon3-3B-Base        & Base & 3B  \\
            & Falcon3-3B-Instruct    & Instruct & 3B  \\
            & Falcon3-7B-Base        & Base & 7B  \\
            & Falcon3-7B-Instruct    & Instruct & 7B  \\
            & Falcon3-10B-Base       & Base & 10B \\
            & Falcon3-10B-Instruct   & Instruct & 10B \\
        \midrule
        \multirow{4}{*}{Llama-3.2}
            & Llama-3.2-1B           & Base & 1B \\
            & Llama-3.2-1B-Instruct  & Instruct & 1B \\
            & Llama-3.2-3B           & Base & 3B \\
            & Llama-3.2-3B-Instruct  & Instruct & 3B \\
        \midrule
        \multirow{4}{*}{Gemma-2}
            & gemma-2-2b             & Base & 2B \\
            & gemma-2-2b-it          & Instruct & 2B \\
            & gemma-2-9b             & Base & 9B \\
            & gemma-2-9b-it          & Instruct & 9B \\
        \midrule
        \multirow{2}{*}{Mistral}
            & Mistral-7B-v0.3        & Base & 7B \\
            & Mistral-7B-Instruct-v0.3 & Instruct & 7B \\
        \midrule
        \multirow{4}{*}{DeepSeek}
            & deepseek-llm-7b-base       & Base & 7B \\
            & deepseek-llm-7b-chat       & Instruct & 7B \\
            & deepseek-math-7b-base      & Base & 7B \\
            & deepseek-math-7b-instruct  & Instruct & 7B \\
        \bottomrule
    \end{tabular}
\end{table*}

\subsection{Benchmark Details}
\label{appendix:benchmark_details}

We provide a complete listing of all textual and multimodal benchmarks used in this study. Textual benchmarks are grouped into six capability dimensions as described in Sec.~\ref{sec:benchmark}, and are summarized in Table~\ref{tab:text_benchmarks}. Multimodal benchmarks are grouped into four capability dimensions as described in Sec.~\ref{sec:benchmark}, and are summarized in Table~\ref{tab:multimodal_benchmarks}.

\begin{table*}[t]
    \centering
    \caption{Textual benchmarks used in this study, grouped by capability dimension. BB: BIG-Bench~\citep{srivastava2022beyond}; BBH: BIG-Bench Hard~\citep{suzgun2022challenging}; MMLU: Massive Multitask Language Understanding~\citep{hendryckstest2021}.}
    \label{tab:text_benchmarks}
    \begin{tabular}{lll}
        \toprule
        \textbf{Category} & \textbf{Benchmark} & \textbf{Source} \\
        \midrule
        Information Extraction
            & CrossNER & \citep{liu2021crossner} \\
        \midrule
        \multirow{6}{*}{Knowledge}
            & hindu\_knowledge          & BB \\
            & mmlu\_stem                & MMLU \\
            & mmlu\_humanities          & MMLU \\
            & mmlu\_other               & MMLU \\
            & dark\_humor\_detection    & BB \\
            & anachronisms              & BB \\
        \midrule
        \multirow{12}{*}{Language}
            & play\_dialog\_same\_or\_different             & BB \\
            & word\_unscrambling                            & BB \\
            & contextual\_parametric\_knowledge\_conflicts  & BB \\
            & wic                                           & \citep{pilehvar2019wic} \\
            & winogrande                                    & \citep{sakaguchi2021winogrande} \\
            & mnist\_ascii                                  & BB \\
            & movie\_dialog\_same\_or\_different            & BB \\
            & ascii\_word\_recognition                      & BB \\
            & disfl\_qa                                     & BB \\
            & word\_sorting\_hard                           & BBH \\
            & winowhy                                       & BB \\
            & tense                                         & BB \\
        \midrule
        \multirow{3}{*}{Math}
            & dyck\_languages\_hard     & BBH \\
            & matrixshapes              & BB \\
            & checkmate\_in\_one        & BB \\
        \midrule
        \multirow{5}{*}{NLI/NLU}
            & unit\_conversion          & BB \\
            & multirc                   & \citep{MultiRC2018} \\
            & symbol\_interpretation    & BB \\
            & undo\_permutation         & BB \\
            & piqa                      & \citep{Bisk2020} \\
        \midrule
        \multirow{13}{*}{Reasoning}
            & boolean\_expressions\_hard    & BBH \\
            & logic\_grid\_puzzle           & BB \\
            & presuppositions\_as\_nli      & BB \\
            & logiqa                        & \citep{liu2020logiqa} \\
            & entailed\_polarity            & BB \\
            & proofwriter\_cwa              & \citep{tafjord2021proofwriter} \\
            & proofwriter\_owa              & \citep{tafjord2021proofwriter} \\
            & mathematical\_induction       & BB \\
            & fantasy\_reasoning            & BB \\
            & analogical\_similarity        & BB \\
            & causal\_judgement\_hard       & BBH \\
            & navigate\_hard                & BBH \\
            & formal\_fallacies\_hard       & BBH \\
        \bottomrule
    \end{tabular}
\end{table*}

\begin{table*}[t]
    \centering
    \caption{Multimodal benchmarks used in this study, grouped by capability dimension.}
    \label{tab:multimodal_benchmarks}
    \begin{tabular}{ll}
        \toprule
        \textbf{Category} & \textbf{Benchmark} \\
        \midrule
        \multirow{13}{*}{General VQA}
            & MMBench\_DEV\_EN\_V11~\citep{liu2024mmbench} \\
            & MMStar~\citep{chen2024we} \\
            & SEEDBench\_IMG~\citep{li2023seed} \\
            & SEEDBench2~\citep{li2023seed2} \\
            & SEEDBench2\_Plus~\citep{li2024seed2plus} \\
            & MME~\citep{fu2026mme} \\
            & TaskMeAnything\_v1\_imageqa\_random~\citep{zhang2024task} \\
            & A-OKVQA~\citep{schwenk2022okvqa} \\
            & RealWorldQA~\citep{grok15} \\
            & HRBench4K~\citep{wang2025divide} \\
            & MMVet~\citep{yu2024mm} \\
            & MME-RealWorld-Lite~\citep{zhang2025mme} \\
            & VStarBench~\citep{vstar} \\
        \midrule
        \multirow{10}{*}{STEM Puzzle}
            & MMSci\_DEV\_MCQ~\citep{li2024mmsci} \\
            & MMMU\_DEV\_VAL~\citep{yue2023mmmu} \\
            & MathVista\_MINI~\citep{lu2024mathvista} \\
            & DynaMath~\citep{zou2024dynamathdynamicvisualbenchmark} \\
            & ScienceQA\_VAL~\citep{lu2022learn} \\
            & AI2D\_TEST~\citep{kembhavi2016diagram} \\
            & MicroVQA~\citep{burgess2025microvqa} \\
            & PathMMU\_VAL~\citep{sun2024pathmmu} \\
            & MathVision~\citep{wang2024measuring} \\
            & MathVerse\_MINI~\citep{zhang2024mathverse} \\
        \midrule
        \multirow{5}{*}{Document Understanding}
            & InfoVQA\_VAL~\citep{mathew2022infographicvqa} \\
            & TableVQABench~\citep{kim2024tablevqa} \\
            & OCRVQA\_TEST~\citep{mishraICDAR19} \\
            & DocVQA\_VAL~\citep{Mathew_2021_WACV} \\
            & GQA\_TestDev\_Balanced~\citep{hudson2018gqa} \\
        \midrule
        \multirow{7}{*}{Alignment}
            & MIA-Bench~\citep{qian2025mia} \\
            & HallusionBench~\citep{Guan_2024_CVPR} \\
            & MMVP~\citep{tong2024eyes} \\
            & LLaVABench~\citep{liu2023llava} \\
            & POPE~\citep{Li-hallucination-2023} \\
            & AesBench\_VAL~\citep{AesBench} \\
            & AMBER~\citep{wang2023llm} \\
        \bottomrule
    \end{tabular}
\end{table*}

\subsection{Benchmark Collection and Filtering}
\label{sec:benchmark_filtering}

We initially collected and evaluated a large pool of over 200 benchmark subsets. The final benchmark list reported in Appendix Tables~\ref{tab:text_benchmarks} contains 40 benchmark entries after a series of filtering and consolidation steps. The difference between the initial collection and the final benchmark set is mainly due to the following processing procedures:

\textbf{(1) Language filtering.}
We exclude non-English benchmarks from the final evaluation set. Although multilingual capabilities are valuable, different model families exhibit substantially different levels of multilingual support, which may introduce additional variance unrelated to the capability transfer studied in this work. Therefore, we focus on English benchmarks to ensure a controlled comparison across model families.

\textbf{(2) Subset merging.}
We merge highly related subsets originating from the same dataset into a single benchmark entry to reduce redundancy and avoid over-weighting datasets with multiple variants. For example, \texttt{logiqa:mrc} and \texttt{logiqa:nli} are aggregated into a unified \texttt{logiqa} benchmark entry.

\textbf{(3) Capability-oriented benchmark filtering.}
We further filter benchmarks that provide limited or redundant signals for constructing a general textual capability representation. Specifically, we remove benchmarks whose performance patterns are highly redundant with existing benchmarks or provide insufficient additional information for distinguishing model capabilities. This step prevents the learned capability representation from being dominated by duplicated signals, ensuring that the final benchmark set provides complementary signals for textual capability modeling.

After these processing steps, we obtain the final set of 40 benchmark entries used for constructing the textual capability representation and fitting the proposed scaling law.

\subsection{Training Details}
\label{training_details}

We implement our method based on the official codebase of
LLaVA-OneVision\footnote{\url{https://github.com/LLaVA-VL/LLaVA-NeXT/tree/main}}.
The training is divided into two stages, with detailed hyperparameter
configurations summarized in Table~\ref{training_hypara}.

\begin{table*}[t]
\centering
\small 
\renewcommand{\arraystretch}{1.15} 
\setlength{\tabcolsep}{8pt} 
\caption{Complete list of transfer coefficients $\lambda_j$ for all textual benchmarks (subsets are merged), ranked by absolute value and categorized into Positive Transfer and Transfer Tax regimes.}
\label{tab:complete_coefficients}
\begin{tabular}{clc @{\hskip 3.5em} clc}
\toprule
\textbf{Rank} & \textbf{Benchmark} & $\lambda_j$ & \textbf{Rank} & \textbf{Benchmark} & $\lambda_j$ \\
\midrule
\multicolumn{3}{c}{\textbf{Positive Transfer ($\lambda_j > 0$)}} & \multicolumn{3}{c}{\textbf{Positive Transfer (Continued)}} \\
\cmidrule(r){1-3} \cmidrule(l){4-6}
1  & \textit{dyck\_languages\_hard} & +0.119 & 21 & \textit{checkmate\_in\_one} & +0.019 \\
2  & \textit{hindu\_knowledge} & +0.108 & 22 & \textit{undo\_permutation} & +0.019 \\
3  & \textit{play\_dialog\_same\_or\_different} & +0.082 & 23 & \textit{mmlu\_other} & +0.016 \\
4  & \textit{word\_unscrambling} & +0.063 & 24 & \textit{causal\_judgement\_hard} & +0.015 \\
5  & \textit{mmlu\_stem} & +0.057 & 25 & \textit{winowhy} & +0.015 \\
6  & \textit{logic\_grid\_puzzle} & +0.050 & 26 & \textit{CrossNER} & +0.014 \\
7  & \textit{matrixshapes} & +0.049 & 27 & \textit{navigate\_hard} & +0.013 \\
8  & \textit{mmlu\_humanities} & +0.046 & 28 & \textit{anachronisms} & +0.009 \\
9  & \textit{wic} & +0.046 & 29 & \textit{tense} & +0.002 \\
\cmidrule(l){4-6}
10 & \textit{winogrande} & +0.045 & \multicolumn{3}{c}{\textbf{Transfer Tax ($\lambda_j < 0$)}} \\
\cmidrule(l){4-6}
11 & \textit{movie\_dialog\_same\_or\_different} & +0.041 & 1  & \textit{boolean\_expressions\_hard} & -0.078 \\
12 & \textit{unit\_conversion} & +0.041 & 2  & \textit{contextual\_param\_knowledge\_conflicts} & -0.048 \\
13 & \textit{presuppositions\_as\_nli} & +0.039 & 3  & \textit{mnist\_ascii} & -0.043 \\
14 & \textit{logiqa} & +0.038 & 4  & \textit{entailed\_polarity} & -0.037 \\
15 & \textit{mathematical\_induction} & +0.036 & 5  & \textit{proofwriter\_cwa} & -0.037 \\
16 & \textit{ascii\_word\_recognition} & +0.035 & 6  & \textit{proofwriter\_owa} & -0.036 \\
17 & \textit{disfl\_qa} & +0.035 & 7  & \textit{multirc} & -0.035 \\
18 & \textit{fantasy\_reasoning} & +0.030 & 8  & \textit{word\_sorting\_hard} & -0.024 \\
19 & \textit{symbol\_interpretation} & +0.030 & 9  & \textit{formal\_fallacies\_hard} & -0.011 \\
20 & \textit{analogical\_similarity} & +0.026 & 10 & \textit{dark\_humor\_detection} & -0.010 \\
   &                                 &        & 11 & \textit{piqa} & -0.000 \\
\bottomrule
\end{tabular}
\end{table*}

\begin{table}[ht]
\centering
\caption{Configuration for training across various stages.}
\label{training_hypara}
\begin{tabular}{@{}clcc@{}}
\toprule
\multicolumn{2}{c}{} & \textbf{Stage-1} & \textbf{Stage-2} \\ 
\midrule
\multirow{2}{*}{\rotatebox[origin=c]{90}{\textit{Vision}}} 
    & \textbf{Resolution} & 384 & 384\footnotesize{$\times$\{(1$\times$1),...,(6$\times$6)\}} \\
    & \#tokens & 729 & Max 10$\times$729 \\ 
\midrule
\rotatebox[origin=c]{90}{\textit{Data}} 
    & \textbf{Samples} & 5M & 12M \\ 
\midrule
\rotatebox[origin=c]{90}{\textit{Model}} 
    & \textbf{Trainable} & Projector & Full Model \\ 
\midrule
\multirow{3}{*}{\rotatebox[origin=c]{90}{\textit{Training}}} 
    & \textbf{Batch Size} & 512 & 512 \\
    & \textbf{LR} & $1\times10^{-3}$ & $1\times10^{-5}$ \\
    & \textbf{Epoch} & 1 & 1 \\ 
\bottomrule
\end{tabular}
\end{table}

\textbf{Model Architecture.}
All models are built upon the LLaVA-OneVision architecture~\citep{li2024llava},  comprising a
vision tower, a projector, and a language tower.
Specifically, the vision tower employs SigLIP~\citep{zhai2023sigmoid} ($\sim$400M parameters) 
to extract visual features from input images. A two-layer MLP with GELU 
activation~\citep{hendrycks2016gaussian,liu2024improved} then serves as 
the projector to map visual features into the language embedding space.

\textbf{Training Data.}
We adopt the Infinity-MM dataset~\citep{gu2024infinity}, which provides
open-source access and sufficient scale for our experiments.
Due to computational constraints, we train on a curated subset rather than
the full corpus. In Stage 1, approximately 5M samples are drawn from a 50\%
random subset of its Stage 1 partition. In Stage 2, we construct
a combined set of approximately 12M samples, comprising the complete
Stages 3--4 data along with ${\sim}3$M samples from its Stage 2 partition.

\textbf{Visual Representations.}
Following~\citet{li2024llavanext-ablations}, we adopt the AnyResMax-9 strategy
(up to 9 sub-image tiles) to balance visual detail and 
computational cost for single-image inputs. 

\section{Methodology for Fitting Multimodal Training Loss}

\label{appendix:method_for_fitting_loss}

To robustly fit the scaling law parameters, we minimize the Huber loss~\citep{huber_robust_1964} between the predicted and observed training loss using the L-BFGS algorithm~\citep{nocedal_updating_1980}:

\begin{align}
    \min_{A, B, E, \alpha, \beta}\quad &\sum_{\text{Runs }i} \text{Huber}_\delta \Big(\log \hat L(N_i, D_i) - \log L_i\Big) \label{eq:huber}
\end{align}

\noindent where  $A, B, \alpha, \beta$ and $E$ are the scaling law parameters and $\delta = 10^{-3}$ is the threshold hyperparameter. 

Given that the objective function for fitting scaling laws is highly non-convex, gradient-based optimizers like L-BFGS are susceptible to local minima. To mitigate this risk, we perform the optimization across multiple independent random initializations. We then select the parameter configuration that yields the lowest Huber loss on the training set as our final fit.

Training loss is recorded at every 10 optimization steps, yielding a dense trajectory for reliable curve fitting. To avoid the highly volatile loss behavior observed in the early phase of training, we exclude data points prior to step 1000 and fit the scaling law exclusively on loss trajectories from step 1000 onward, where the training dynamics have stabilized and the loss curve follows a smoother convergence trend.

Furthermore, prior to fitting, we apply a centered rolling average with a window size of 5 to smooth the raw training loss curves:

\begin{equation}
    \tilde{L}_t = \frac{1}{|\mathcal{W}_t|} \sum_{i \in \mathcal{W}_t} L_i
\end{equation}

\noindent where $\mathcal{W}_t$ denotes the set of indices within the window centered at step $t$, and $|\mathcal{W}_t|$ accounts for boundary effects (i.e., \texttt{min\_periods=1}). This smoothing step attenuates high-frequency noise in the loss trajectory while preserving the underlying convergence trend, leading to more stable and generalizable fits.

\section{Optimization Algorithm for the Performance Predictor}
\label{sec:appendix_algorithm}

We detail the optimization procedure for the multimodal performance predictor (Eq.~\ref{eq:multimodal_scaling}), as outlined in Algorithm~\ref{alg:fit_trajectory_law}. The objective is to jointly learn the capability aggregation weights $\mathbf{w}$ and the scaling coefficients $\{\hat{A}, B_0, B_m, P_0\}$ from $M$ LLM--VLM training trajectories.

\definecolor{blue(ncs)}{rgb}{0.0, 0.53, 0.74}

\SetKwComment{Comment}{}{\ }
\newcommand\mytcp[1]{\Comment{\hfill\textnormal{\textcolor{blue(ncs)}{$\triangleright$ #1}}}}
\let\tcp\mytcp
\newcommand\mytcc[1]{\Comment{\textnormal{\textcolor{blue(ncs)}{/* #1 \hfill */}}}}
\let\tcc\mytcc
\SetKwInput{KwParameters}{Args}

\begin{algorithm*}[t]
\SetAlgoLined

\KwParameters{
    number of LLM--VLM pairs $M$,
    number of text benchmarks $T$,
    number of principal components $K$,
    number of checkpoints $N_{ckpt}$,
    absorption penalty weight $\lambda$
}

\KwIn{
    text benchmark matrix $\mathbf{X} \in \mathbb{R}^{T \times M}$,\quad
    multimodal data scales $D_\text{mm} \in \mathbb{R}^{N_{ckpt}}$,\quad
    multimodal accuracy trajectories $P \in \mathbb{R}^{M \times N_{ckpt}}$
}

\KwResult{
    fitted performance predictor $\mathcal{F}$,\quad
    optimal parameters $\hat{A}^*, B_0^*, B_m^*, P_0^*, \mathbf{w}^*$
}

~\\
\tcc{Step 1: Extract latent capability representation via PCA (cf.\ Eq.~\ref{eq:obs_scaling})}

$\boldsymbol{\mu}
\leftarrow
\text{Mean}(\mathbf{X},\,\text{axis}=1)$
\tcp{Compute benchmark-wise means across training LLM backbones}

$\mathbf{X}_c
\leftarrow
\mathbf{X}-\boldsymbol{\mu}$
\tcp{Center each benchmark dimension using training-set statistics}

$\boldsymbol{\Gamma}, \mathbf{S}
\leftarrow
\text{PCA}(\mathbf{X}_c,\,K)$
\tcp{Fit PC loading vectors $\boldsymbol{\Gamma}\in\mathbb{R}^{K\times T}$ and extract capability matrix $\mathbf{S}=\boldsymbol{\Gamma}\mathbf{X}_c\in\mathbb{R}^{K\times M}$}

\tcp{Select $K$ as the minimum number of components explaining $\geq95\%$ of total variance}

~\\
\tcc{Step 2: Parameterize scalar capability score and predicted trajectory (cf.\ Eq.~\ref{eq:capability_score}--\ref{eq:multimodal_scaling})}

$S_m(\mathbf{w})
\leftarrow
\mathbf{w}^{\top}\mathbf{S}_m,
\quad \forall m$
\tcp{Aggregate latent capability with weights $\mathbf{w}\in\mathbb{R}^{K}$}

$\hat{B}(\mathbf{w})
\leftarrow
B_0-B_m\cdot S_m(\mathbf{w})$
\tcp{Compute model-specific data absorption rate}

$P_{\mathrm{pred}}(m,t;\theta)
\leftarrow
\hat{A}\cdot S_m(\mathbf{w})
+
\hat{B}(\mathbf{w})\cdot\ln D_\text{mm}^{(t)}
+
P_0$
\tcp{Predicted accuracy; $\theta=\{\hat{A},B_0,B_m,P_0,\mathbf{w}\}$}

~\\
\tcc{Step 3: Define objective function with robust loss and physical constraints}

$\mathcal{L}_{\mathrm{Huber}}(\theta)
\leftarrow
\displaystyle
\sum_{\substack{(m,t):\\P_{m,t}\neq\mathrm{NaN}}}
\mathrm{Huber}
\left(
P_{\mathrm{pred}}(m,t;\theta)-P_{m,t}
\right)$
\tcp{Huber loss over valid observations}

$\mathcal{L}_{\mathrm{abs}}(\theta)
\leftarrow
\lambda
\displaystyle
\sum_{m=1}^{M}
\left[
\min
\left(
0,\,
B_0-B_m\cdot S_m(\mathbf{w})
\right)
\right]^2$
\tcp{Soft penalty enforcing non-negative absorption rate}

$\mathcal{L}(\theta)
\leftarrow
\mathcal{L}_{\mathrm{Huber}}(\theta)
+
\mathcal{L}_{\mathrm{abs}}(\theta)$
\tcp{Full optimization objective}

~\\
\tcc{Step 4: Two-stage optimization and capability normalization}

$\theta_{\mathrm{init}}
\leftarrow
\mathrm{DifferentialEvolution}
(\mathcal{L}(\theta))$
\tcp{Gradient-free global search to escape local optima}

$\theta^*
\leftarrow
\mathrm{L\mbox{-}BFGS\mbox{-}B}
(\mathcal{L}(\theta),\theta_{\mathrm{init}})$
\tcp{Gradient-based refinement without explicit unit-norm constraint on $\mathbf{w}$}

$\hat{A}^*,B_0^*,B_m^*,P_0^*,\mathbf{w}^*
\leftarrow
\theta^*$

$r
\leftarrow
\|\mathbf{w}^*\|_2$
\tcp{Compute scale factor of optimized capability direction}

$\mathbf{w}^*
\leftarrow
\frac{\mathbf{w}^*}{r}$
\tcp{Normalize capability direction to unit norm}

$\hat{A}^* \leftarrow \hat{A}^*\cdot r,\quad
B_m^* \leftarrow B_m^*\cdot r$
\tcp{Preserve predictions after normalizing $\mathbf{w}$}

~\\
\tcc{Step 5: Inference --- predict trajectory of any unseen model}

\Return{
$
\mathcal{F}:
(X_{\mathrm{new}},D_\text{mm})
\mapsto
\hat{A}^*S_{\mathrm{new}}
+
(B_0^*-B_m^*S_{\mathrm{new}})
\ln D_\text{mm}
+
P_0^*
$
}

\tcp{
where
$
S_{\mathrm{new}}
=
\mathbf{w}^{*\top}
\boldsymbol{\Gamma}
(X_{\mathrm{new}}-\boldsymbol{\mu})
$;
$\boldsymbol{\mu}$ is computed from training LLM backbones
}

\caption{Fitting the Capability-Driven Multimodal Performance Predictor}
\label{alg:fit_trajectory_law}

\end{algorithm*}

\subsection*{Algorithm Walkthrough}

\paragraph{Step 1 --- Latent Capability Extraction.}
To prevent overfitting on high-dimensional text benchmarks~\citep{ruan2024observational}, we apply PCA to the benchmark-model matrix $\mathbf{X}$. We extract a low-dimensional latent representation $\mathbf{S}_m \in \mathbb{R}^K$ that captures at least 95\% of the total variance.

\paragraph{Step 2 --- Joint Trajectory Parameterization.}
The scalar capability score $S_m = \mathbf{w}^\top \mathbf{S}_m$ jointly drives the \emph{transfer} term ($\hat{A} \cdot S_m$) and the \emph{absorption} term ($\hat{B} \cdot \ln D_\text{mm}$). Optimizing $\mathbf{w}$ and the scaling parameters end-to-end ensures that $\mathbf{w}$ captures a holistic textual capability measure governing the entire multimodal learning dynamics. During optimization, the capability direction $w$ is treated as an unconstrained parameter. After convergence, $\mathbf{w}$ is normalized to satisfy the unit-norm constraint, and the normalization factor is absorbed into both $\hat{A}$ and $B_m$.

\paragraph{Step 3 --- Robust Objective with Constraints.}
We employ the Huber loss for robustness against noisy evaluations and missing checkpoints. A soft penalty enforces $\hat{B} \geq 0$, satisfying the physical constraint that additional multimodal data should not degrade performance.

\paragraph{Step 4 --- Two-Stage Optimization.}
The bilinear interaction ($B_m \cdot S_m \cdot \ln D_\text{mm}$) renders the objective non-convex. Therefore, we utilize Differential Evolution~\citep{storn1995differrential} for global search, followed by L-BFGS-B~\citep{nocedal_updating_1980} for precise local refinement.

\paragraph{Step 5 --- Zero-Shot Trajectory Prediction.}
Once fitted, the framework predicts complete multimodal learning trajectories of unseen backbones strictly from their text benchmark scores, enabling zero-cost, optimal LLM selection prior to expensive multimodal training runs.

\section{Comparison with Alternative Predictors}
\label{sec:alternative_predictors}

\begin{figure*}[t]
    \centering

    \begin{subfigure}[t]{0.48\textwidth}
        \centering
        \includegraphics[width=\linewidth]{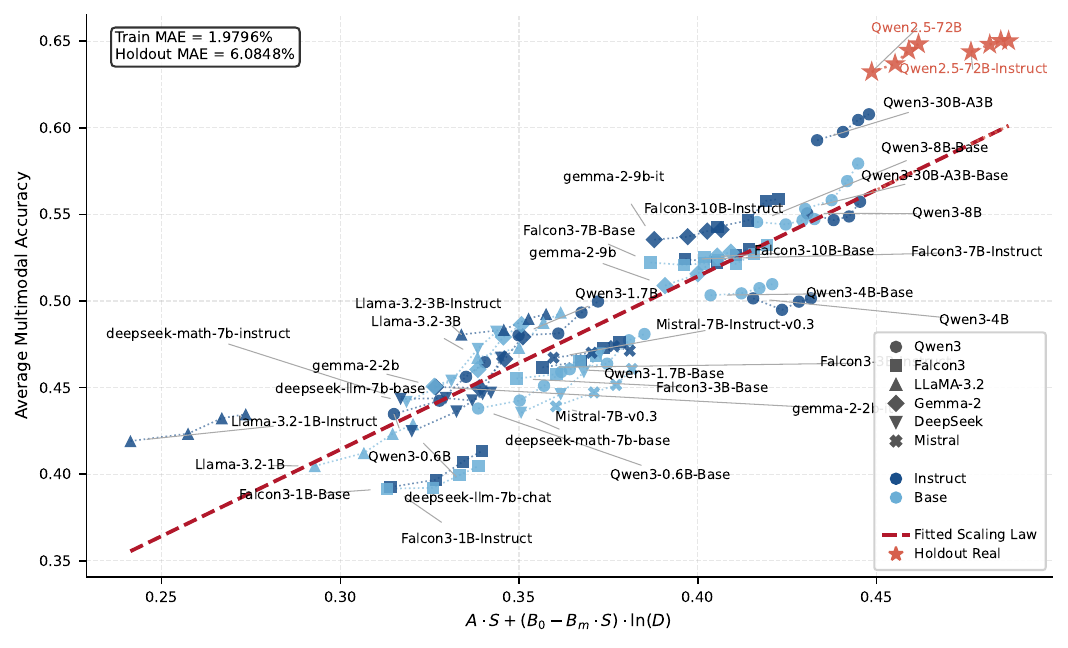}
        \caption{Average textual benchmark score}
        \label{fig:avgscore_fitting}
    \end{subfigure}
    \hfill
    \begin{subfigure}[t]{0.48\textwidth}
        \centering
        \includegraphics[width=\linewidth]{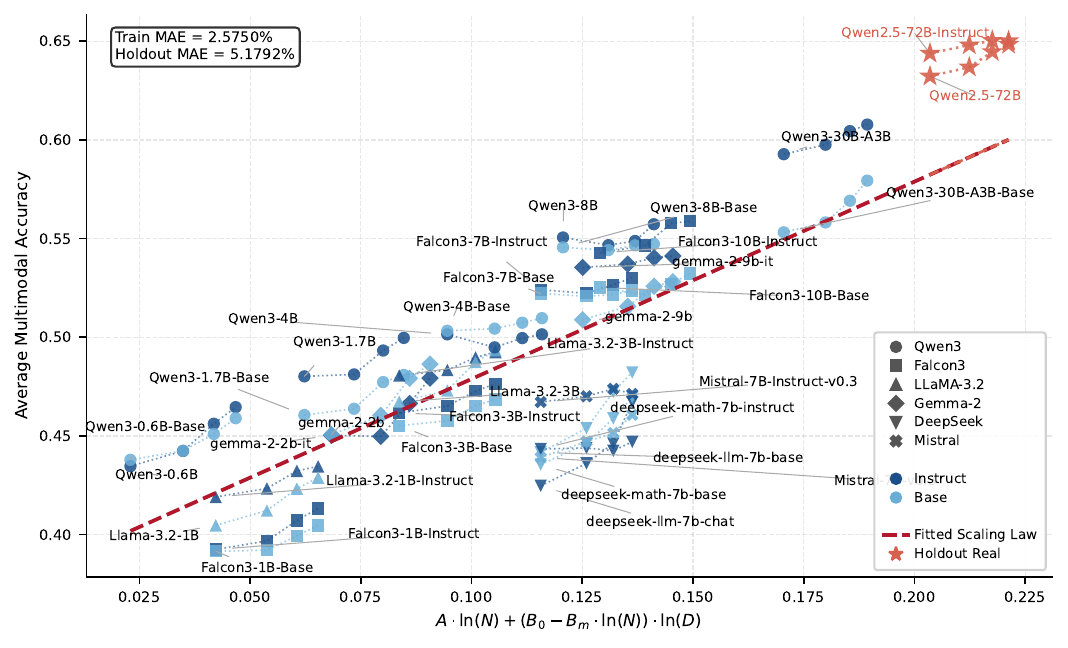}
        \caption{Model parameter count ($\log N$)}
        \label{fig:param_fitting}
    \end{subfigure}

    \caption{
    Comparison of two alternative predictors for VLM performance. Both average textual benchmark score and model parameter count provide coarse fitting trends but exhibit substantially weaker extrapolation behavior compared with the proposed capability-driven formulation shown in Figure~\ref{fig:acc_fitting}.
    }
    \label{fig:baseline_compare}
\end{figure*}

To further contextualize the proposed capability-driven scaling law, we compare it against two direct baselines: (1) average textual benchmark score and (2) model parameter count ($\log N$). Figure~\ref{fig:baseline_compare} visualizes the corresponding scaling fits and 72B extrapolation behavior, while Figure~\ref{fig:acc_fitting} presents the proposed capability-driven formulation.

Compared with both baselines, our method achieves substantially lower fitting and extrapolation errors. In particular, average textual benchmark score and parameter count exhibit noticeably weaker extrapolation behavior at larger scales, whereas the proposed capability-driven representation maintains stable predictive accuracy across both fitting and held-out prediction regimes. Notably, parameter count alone cannot distinguish between models with substantially different textual capabilities but similar parameter scales, leading to nearly identical VLM performance predictions for equally-sized backbones.

\section{Robustness to Highly Correlated Text Benchmarks}
\label{sec:correlation_robustness}

To evaluate whether the proposed scaling law depends disproportionately on a small number of highly correlated textual benchmarks, we analyze the relationship between text and multimodal benchmarks using Spearman rank correlation across all benchmark pairs. We observe that several textual benchmarks, including \texttt{mmlu\_stem}, \texttt{matrixshapes}, and \texttt{formal\_fallacies\_hard}, exhibit particularly high correlation with the multimodal benchmark DynaMath (maximum $\rho > 0.9$).

\begin{figure}[htbp]
    \centering
    \includegraphics[width=\linewidth]{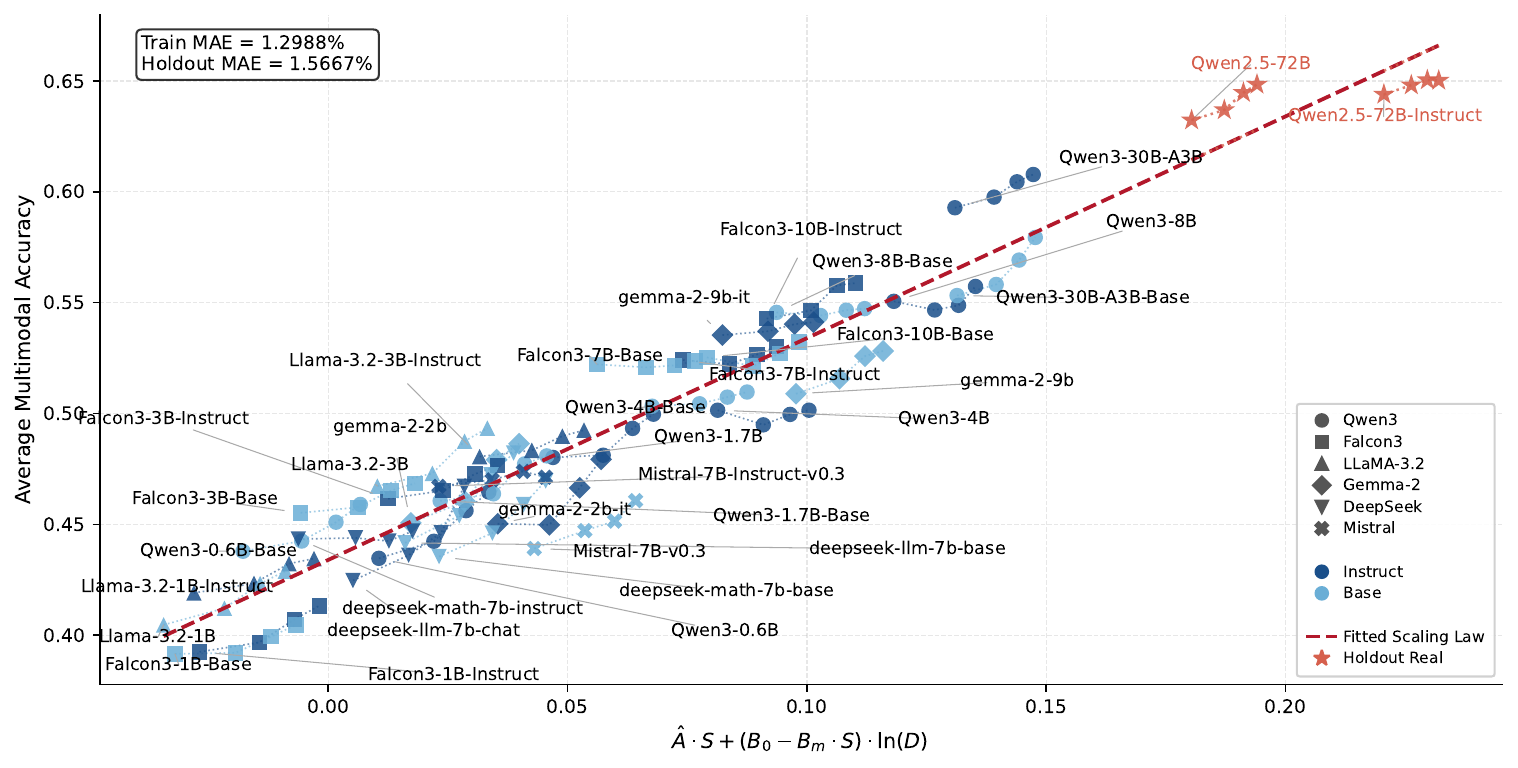}
    \caption{
    Robustness analysis after removing highly correlated textual benchmarks.
    All textual benchmarks whose maximum Spearman correlation with any multimodal benchmark exceeds 0.9 are excluded before refitting the capability-driven scaling law.
    The resulting model maintains nearly identical fitting error and only moderately increased holdout prediction error, indicating that the proposed framework does not rely on a small set of highly correlated benchmarks.
    }
    \label{fig:correlation_filter}
\end{figure}

We therefore remove all textual benchmarks whose maximum correlation with any multimodal benchmark exceeds 0.9, and refit the capability-driven scaling model using only the remaining benchmarks. As shown in Figure~\ref{fig:correlation_filter}, the fitting MAE changes only marginally from 1.287\% to 1.299\%, while the holdout MAE increases moderately from 1.225\% to 1.567\%.

These results suggest that the predictive performance of the proposed framework does not depend on a small set of highly correlated benchmarks, but instead emerges from broader transferable textual capability signals.

\section{Category-Specific Multimodal Scaling Dynamics}
\label{sec:category_scaling}

Our main analysis uses the average score across all multimodal benchmarks as the prediction target, providing a broad measure of overall VLM capability. However, different multimodal tasks may rely on textual priors and multimodal training data to different degrees. To examine whether such task-specific differences are captured by our framework, we further decompose the multimodal evaluation suite according to the taxonomy in Table~\ref{tab:multimodal_benchmarks} and independently fit Eq.~\ref{eq:multimodal_scaling} on four capability categories: \textbf{General VQA}, \textbf{Document Understanding}, \textbf{STEM Puzzle}, and \textbf{Alignment}. For each category, the target performance is computed as the average score over its constituent benchmarks, while the textual capability representation and fitting procedure remain unchanged.

The resulting parameters are summarized in Table~\ref{tab:category_fitting}. The fitted coefficients vary substantially across capability categories, revealing heterogeneous transfer and absorption dynamics. General VQA exhibits the largest transfer coefficient ($\hat{A}=0.322$), indicating the strongest fitted association between textual backbone capability and downstream performance among the four categories. In comparison, Document Understanding and STEM Puzzle yield smaller transfer coefficients ($\hat{A}=0.228$ and $0.212$, respectively), suggesting that the contribution associated with textual capability differs considerably across multimodal capability groups.

The data-scaling terms exhibit similarly distinct behaviors. Document Understanding has the largest average absorption rate ($\bar{B}=2.601\times10^{-2}$), suggesting comparatively strong gains from additional multimodal training data. STEM Puzzle, in contrast, has a very small capability-dependent absorption decay ($B_m=0.098\times10^{-2}$), indicating that its data-scaling rate remains relatively stable across backbones with different textual capability scores. Alignment displays the opposite pattern: although its average absorption rate is relatively low ($\bar{B}=0.728\times10^{-2}$), it has the largest absorption decay coefficient ($B_m=2.309\times10^{-2}$), showing the strongest dependence of data-scaling efficiency on the textual capability of the backbone.

Despite these category-specific differences, the proposed scaling law maintains consistently low prediction errors, with MAEs ranging from $0.996\%$ to $1.974\%$ across all four categories. This suggests that the capability-driven formulation generalizes beyond aggregate multimodal performance, while revealing distinct category-specific transfer--absorption profiles.

\begin{table}[t]
\centering
\caption{
Category-specific fitting of the Capability-Driven Multimodal Scaling Law.
Multimodal benchmarks are grouped according to the taxonomy in Table~\ref{tab:multimodal_benchmarks}, and Eq.~\ref{eq:multimodal_scaling} is fitted independently for each category.
For consistency, $B_0$, $B_m$, and $\bar{B}$ are scaled by $10^2$.
}
\label{tab:category_fitting}
\resizebox{0.65\linewidth}{!}{
\begin{tabular}{lccccc}
\toprule
Category & $\hat{A}$ & $B_0$ ($10^{-2}$) & $B_m$ ($10^{-2}$) & $\bar{B}$ ($10^{-2}$) & MAE (\%) \\
\midrule
General VQA            & 0.322 & 1.502 & 1.056 & 1.501 & 1.552 \\
Document Understanding & 0.228 & 2.601 & 0.917 & 2.601 & 1.974 \\
STEM Puzzle            & 0.212 & 1.665 & 0.098 & 1.665 & 1.583 \\
Alignment              & 0.245 & 0.728 & 2.309 & 0.728 & 0.996 \\
\bottomrule
\end{tabular}
}
\end{table}

\section{Robustness of Transfer Coefficients}
\label{sec:transfer_robustness}

The transfer coefficients analyzed in Sec.~\ref{sec:analysis_tax} are estimated from a particular collection of textual benchmarks and LLM backbones. We therefore examine whether the identified positive-transfer and transfer-tax patterns remain stable under perturbations to the benchmark composition and model-family mixture. Specifically, we refit the capability-driven predictor under six complementary robustness settings: randomly removing approximately $12.5\%$ of the textual benchmarks, performing four leave-one-family-out fits by excluding Qwen3, DeepSeek, Falcon3, and Gemma-2 respectively, and constructing a contamination-resistant setting by removing textual benchmarks exhibiting exceptionally high correlations with individual multimodal benchmarks.

For each textual benchmark, we record the signs of its fitted transfer coefficients across the available settings. We classify a benchmark as \emph{stable positive transfer} if $\lambda_j>0$ in at least five settings, as a \emph{stable transfer tax} if $\lambda_j<0$ in at least five settings, and as \emph{unstable} otherwise. When a benchmark is excluded by the contamination-resistant filtering procedure, statistics are computed over the remaining available fits. Table~\ref{tab:transfer_tax_robustness} reports the mean and standard deviation of $\lambda_j$, together with the numbers of positive and negative coefficients across robustness settings.

Overall, the coefficient signs remain largely stable across these perturbations. Among the 40 textual benchmarks, 25 ($62.5\%$) are identified as stable positive-transfer benchmarks, 6 ($15.0\%$) as stable transfer-tax benchmarks, and 9 ($22.5\%$) as unstable. These results are broadly consistent with the capability-level interpretation in Sec.~\ref{sec:analysis_tax}: positive-transfer benchmarks such as \texttt{matrixshapes}, \texttt{dyck\_languages\_hard}, and \texttt{mmlu\_stem} remain consistently positive, while transfer-tax benchmarks such as \texttt{multirc}, \texttt{entailed\_polarity}, and \texttt{contextual\_parametric\_knowledge\_conflicts} remain consistently negative across different robustness settings. Meanwhile, near-neutral benchmarks such as \texttt{piqa} exhibit sign changes across settings, consistent with our interpretation that such coefficients may be sensitive to saturation or dimensional orthogonality. Overall, more than three quarters of the benchmarks preserve a consistent transfer direction, supporting the robustness of the positive-transfer and transfer-tax patterns observed in the main analysis.

\begin{table*}[t]
\centering
\caption{
Robustness of benchmark-level transfer coefficients across six complementary fitting settings.
Benchmarks are classified as \emph{Stable Positive Transfer} if $\lambda_j>0$ in at least five settings,
as \emph{Stable Transfer Tax} if $\lambda_j<0$ in at least five settings, and as \emph{Unstable} otherwise.
For benchmarks excluded by the contamination-resistant filtering procedure, statistics are computed over the remaining available fits.
Benchmarks within each group are ordered by the absolute value of their mean transfer coefficient.
}
\label{tab:transfer_tax_robustness}
\resizebox{\textwidth}{!}{
\begin{tabular}{lrrrr@{\qquad}lrrrr}
\midrule
\multicolumn{10}{c}{\textbf{Stable Positive Transfer}} \\
\midrule
Benchmark & Mean $\lambda$ & Std. $\lambda$ & Pos. & Neg. &
Benchmark & Mean $\lambda$ & Std. $\lambda$ & Pos. & Neg. \\
\midrule
\texttt{dyck\_languages\_hard}              & 0.374 & 0.156 & 6 & 0 &
\texttt{symbol\_interpretation}             & 0.118 & 0.038 & 6 & 0 \\
\texttt{matrixshapes}                       & 0.341 & 0.106 & 5 & 0 &
\texttt{disfl\_qa}                          & 0.116 & 0.048 & 6 & 0 \\
\texttt{hindu\_knowledge}                   & 0.282 & 0.091 & 6 & 0 &
\texttt{wic}                                & 0.116 & 0.109 & 5 & 1 \\
\texttt{word\_unscrambling}                 & 0.261 & 0.052 & 6 & 0 &
\texttt{mmlu\_humanities}                   & 0.106 & 0.062 & 5 & 1 \\
\texttt{winogrande}                         & 0.217 & 0.097 & 5 & 0 &
\texttt{winowhy}                            & 0.105 & 0.144 & 5 & 1 \\
\texttt{play\_dialog\_same\_or\_different}  & 0.189 & 0.096 & 6 & 0 &
\texttt{anachronisms}                       & 0.097 & 0.074 & 5 & 1 \\
\texttt{mmlu\_stem}                         & 0.175 & 0.020 & 5 & 0 &
\texttt{fantasy\_reasoning}                 & 0.087 & 0.074 & 5 & 1 \\
\texttt{logiqa}                             & 0.144 & 0.061 & 6 & 0 &
\texttt{movie\_dialog\_same\_or\_different} & 0.087 & 0.060 & 6 & 0 \\
\texttt{unit\_conversion}                   & 0.143 & 0.061 & 6 & 0 &
\texttt{undo\_permutation}                  & 0.083 & 0.041 & 6 & 0 \\
\texttt{presuppositions\_as\_nli}           & 0.137 & 0.046 & 6 & 0 &
\texttt{navigate\_hard}                     & 0.079 & 0.049 & 5 & 1 \\
\texttt{logic\_grid\_puzzle}                & 0.124 & 0.093 & 6 & 0 &
\texttt{word\_sorting\_hard}                & 0.077 & 0.099 & 5 & 1 \\
\texttt{ascii\_word\_recognition}           & 0.123 & 0.028 & 6 & 0 &
\texttt{mmlu\_other}                        & 0.066 & 0.023 & 6 & 0 \\
                                            &       &       &   &   &
\texttt{causal\_judgement\_hard}            & 0.048 & 0.035 & 6 & 0 \\
\midrule
\multicolumn{10}{c}{\textbf{Stable Transfer Tax}} \\
\midrule
\texttt{multirc}                            & -0.119 & 0.056 & 0 & 6 &
\texttt{contextual\_parametric\_knowledge\_conflicts}
                                            & -0.099 & 0.060 & 0 & 6 \\
\texttt{proofwriter\_cwa}                   & -0.115 & 0.092 & 1 & 5 &
\texttt{proofwriter\_owa}                   & -0.089 & 0.082 & 1 & 5 \\
\texttt{entailed\_polarity}                 & -0.107 & 0.041 & 0 & 6 &
\texttt{dark\_humor\_detection}             & -0.062 & 0.069 & 1 & 5 \\
\midrule
\multicolumn{10}{c}{\textbf{Unstable}} \\
\midrule
\texttt{mnist\_ascii}                       & -0.077 & 0.121 & 1 & 4 &
\texttt{CrossNER}                           &  0.036 & 0.067 & 3 & 3 \\
\texttt{analogical\_similarity}             &  0.075 & 0.060 & 4 & 1 &
\texttt{formal\_fallacies\_hard}            &  0.022 & 0.030 & 3 & 2 \\
\texttt{checkmate\_in\_one}                 &  0.047 & 0.049 & 4 & 1 &
\texttt{tense}                              &  0.021 & 0.045 & 4 & 2 \\
\texttt{mathematical\_induction}            & -0.038 & 0.210 & 3 & 2 &
\texttt{boolean\_expressions\_hard}         &  0.011 & 0.195 & 4 & 2 \\
\texttt{piqa}                               &  0.010 & 0.085 & 2 & 4 &
                                            &        &       &   &   \\
\bottomrule
\end{tabular}
}
\end{table*}

\section{Pair-wise Crossover Analysis between Base and Instruct Variants}
\label{sec:crossover_analysis}

To further analyze the scaling behavior between Base and Instruct variants, we perform a pair-wise scaling analysis on representative matched Base--Instruct model pairs. For each pair, we independently fit the multimodal scaling trajectories of the Base and Instruct variants using a log-linear scaling function:

\begin{equation}
    L(D)=\alpha+\beta\ln(D),
\end{equation}

where $L(D)$ denotes the multimodal benchmark performance and $D$ represents the amount of multimodal training data (in billion tokens). The coefficient $\beta$ captures the performance improvement rate as multimodal data increases.

For each Base--Instruct pair, we obtain the corresponding scaling coefficients $(\alpha_{\rm base},\beta_{\rm base})$ and $(\alpha_{\rm chat},\beta_{\rm chat})$, and define

\begin{equation}
    \Delta\beta=\beta_{\rm base}-\beta_{\rm chat}.
\end{equation}

A positive $\Delta\beta$ indicates that the Base variant improves faster as multimodal data increases. We further estimate the crossover data scale $D^*$, where the two fitted scaling trajectories achieve the same performance, by solving

\begin{equation}
    \alpha_{\rm base}+\beta_{\rm base}\ln(D^*)
    =
    \alpha_{\rm chat}+\beta_{\rm chat}\ln(D^*),
\end{equation}

which gives

\begin{equation}
    \ln(D^*)=
    \frac{\alpha_{\rm chat}-\alpha_{\rm base}}
    {\beta_{\rm base}-\beta_{\rm chat}}.
\end{equation}

Representative results are summarized in Table~\ref{tab:crossover_analysis}.

\begin{table}[t]
\centering
\caption{Representative pair-wise crossover analysis between Base and Instruct variants. $\Delta\beta=\beta_{\rm base}-\beta_{\rm chat}$ measures the relative multimodal scaling slope of Base variants, and $D^*$ denotes the estimated crossover point in billion multimodal training tokens.}
\label{tab:crossover_analysis}
\small
\begin{tabular}{llccc}
\toprule
Family & Backbone & $\Delta\beta$ & $D^*$ (B) & Interpretation \\
\midrule
LLaMA-3.2 & Llama-3.2-1B & +0.828 & 85.2 & Crossover \\
Gemma-2 & Gemma-2-9B & +0.903 & 234.5 & Crossover \\
Qwen3 & Qwen3-30B-A3B & +1.117 & 587.8 & Large-scale crossover \\
Qwen3 & Qwen3-4B & +0.084 & -- & Base-favorable \\
Falcon3 & Falcon3-3B & +0.024 & $7.6\times10^{13}$ & Weak / unstable \\
Falcon3 & Falcon3-10B & -0.098 & -- & Chat-favorable \\
\bottomrule
\end{tabular}
\end{table}

The pair-wise results reveal substantial variation across model families. Llama-3.2-1B and Gemma-2-9B exhibit crossover points at approximately $85$B and $235$B multimodal tokens, respectively, while Qwen3-30B-A3B crosses over at a larger scale of approximately $588$B tokens. Qwen3-4B is already Base-favorable over the considered data range. In contrast, Falcon3 exhibits much weaker Base--Instruct differentiation: Falcon3-3B has nearly identical fitted slopes ($\Delta\beta=0.024$), yielding an extremely distant and unstable crossover, whereas Falcon3-10B has a negative $\Delta\beta$ and exhibits Chat-favorable scaling behavior.

Across all model pairs with positive $\Delta\beta$, the median crossover occurs at $\ln(D^*)=4.95$, corresponding to approximately $141$B multimodal training tokens. Notably, this scale falls within the training budgets of modern large-scale VLMs. For example, Qwen2.5-VL is trained on trillions of tokens ($4.1$T tokens in total), substantially exceeding this median crossover scale.

\end{document}